\documentclass[letterpaper]{article} % DO NOT CHANGE THIS
\usepackage{aaai25}  % DO NOT CHANGE THIS
\usepackage{times}  % DO NOT CHANGE THIS
\usepackage{helvet}  % DO NOT CHANGE THIS
\usepackage{courier}  % DO NOT CHANGE THIS
\usepackage[hyphens]{url}  % DO NOT CHANGE THIS
\usepackage{graphicx} % DO NOT CHANGE THIS
\usepackage{natbib}  % DO NOT CHANGE THIS AND DO NOT ADD ANY OPTIONS TO IT
\usepackage{caption} % DO NOT CHANGE THIS AND DO NOT ADD ANY OPTIONS TO IT
\usepackage{amsmath}
\usepackage{amsfonts}
\usepackage{graphicx}
\usepackage{subfigure}
\usepackage{algorithm}
\usepackage{algorithmic}
\usepackage{newfloat}
\usepackage{listings}
\DeclareCaptionStyle{ruled}{labelfont=normalfont,labelsep=colon,strut=off} % DO NOT CHANGE THIS
\floatstyle{ruled}
\newfloat{listing}{tb}{lst}{}
\floatname{listing}{Listing}
\newcommand{\loss}{\mathcal{L}}

\usepackage{amsthm}
\newtheorem{theorem}{Theorem}

\newtheorem{corollary}[theorem]{Corollary}

\usepackage[toc,page]{appendix}

\title{Convergence Analysis of Federated Learning Methods \\ Using Backward Error Analysis}
\author{
    Jinwoo Lim\textsuperscript{\rm 1},
    Suhyun Kim\textsuperscript{\rm 2}\footnote{Co-corresponding authors.},
    Soo-Mook Moon\textsuperscript{\rm 1}$^*$
}
\affiliations{
    \textsuperscript{\rm 1}Seoul National University\\
    \textsuperscript{\rm 2}Korea Institute of Science and Technology\\
    jinwoolim8180@snu.ac.kr, dr.suhyun.kim@gmail.com, smoon@snu.ac.kr
}

\usepackage{bibentry}
\begin{document}

\maketitle

\begin{abstract}
  % Backward error analysis is a powerful technique that can check how much the path of the gradient flow is modified under the influence of a finite learning rate or an optimization method. Through this technique, it is possible to find a modified loss function that updates of the parameter truly follows, and the additional loss term in the modified loss is called \textit{implicit regularizer}. This paper seeks implicit regularizers of various federated learning algorithms and attempts to understand the implicit bias and limitations of those methods. We prove that the implicit regularizer for FedAvg disperses the gradient of each client from the average gradient, increasing the gradient variance. We then empirically show that the implicit regularizer of FedAvg can hamper convergence. 
  % Moreover, our analysis obtains implicit regularizers of 
  % FedSAM and SCAFFOLD and explains the convergence behavior of those federated learning methods. We prove that FedSAM can only partially remove the bias by the first-order term in implicit regularizer of FedAvg. We also prove that SCAFFOLD can be better than FedSAM since it can completely remove the bias by the first-order term, but shows its limitation in the second-order term of the implicit regularizer.

  %Backward error analysis is a powerful technique that can check how much the path of the gradient flow is modified under the influence of a finite learning rate or an optimization method. Using this technique, it is possible to find a modified loss function that the updates of the parameters really follow. 

  Backward error analysis allows finding a modified loss function, which the parameter updates really follow under the influence of an optimization method. 
  The additional loss terms included in this modified function is called \textit{implicit regularizer}. In this paper, we attempt to find the implicit regularizer for various federated learning algorithms on non-IID data distribution, and explain why each method shows different convergence behavior. We first show that the implicit regularizer of FedAvg disperses the gradient of each client from the average gradient, thus increasing the gradient variance. We also empirically show that the implicit regularizer hampers its convergence. Similarly, we compute the implicit regularizers of FedSAM and SCAFFOLD, and explain why they converge better. While existing convergence analyses focus on pointing out the advantages of FedSAM and SCAFFOLD, our approach can explain their limitations in complex non-convex settings. 
  %Unlike most existing convergence analysis that mainly focus on the advantages of FedSAM and SCAFFOLD, the implicit regularizer can explain their limitations as well, even in a more complex non-convex setting. 
  %For example, we can prove that FedSAM can partially remove the bias for the first-order term of the implicit regularizer of FedAvg, while SCAFFOLD can completely remove it for the first-order term, but not for the second-order term. 
  In specific, we demonstrate that FedSAM can partially remove the bias in the first-order term of the implicit regularizer in FedAvg, whereas SCAFFOLD can fully eliminate the bias in the first-order term, but not in the second-order term.
  Consequently, the implicit regularizer can provide a useful insight on the convergence behavior of federated learning from a different theoretical perspective.
\end{abstract}

\section{Introduction}

Federated learning is a distributed learning technique in which a central server builds a global model by repeatedly aggregating the parameters of local models that clients update and upload. The most popular algorithm is FedAvg \citep{mcmahan2017communication}, which aggregates the parameters simply by averaging them. 
Despite its privacy advantage, FedAvg suffers from lower accuracy and slower convergence % than centralized learning 
due to a misalignment between local and global objectives, especially on non-IID data distribution \citep{zhao2018noniid, karimireddy2020scaffold}. 
Many algorithms, such as FedSAM \citep{caldarola2022fedasam, qu2022generalized} and SCAFFOLD \cite{karimireddy2020scaffold}, have been proposed to reduce the drift in local updates and enhance performance. 
%To reduce this {\em drift} of local updates, many algorithms such as {\em FedSAM} \citep{caldarola2022fedasam, qu2022generalized} or {\em SCAFFOLD} \cite{karimireddy2020scaffold} have been proposed for a better performance. 
Also, there have been many researches to understand the difference of the convergence behavior with an analysis from the convex optimization perspectives \citep{wang2019adaptive, yu2019parallel, li2020convergence, khaled2020tighter, haddadpour2019convergence, karimireddy2020scaffold, qu2022generalized}. However, it is still challenging to make a tight analysis on non-IID data. This paper attempts to analyze the convergence behavior of a few federated learning algorithms from a different theoretical perspective, \textit{implicit regularization}.
%On the other hand, in this paper, we borrow the concept of \textit{implicit regularization} to intuitively investigate the strengths and limitations of various federated learning methods including variance reduction methods as FedSAM, and SCAFFOLD.

The concept of implicit regularization was originally introduced to explain the generalization behavior of centralized \textit{gradient descent} (GD) and \textit{stochastic gradient descent} (SGD) \citep{barrett2021implicit, smith2021on}; unless otherwise stated, GD and SGD are for centralized in this paper. Their approach, inspired by \textit{backward error analysis} \citep{hairer2006geometric}, is to analyze the path on which discrete parameter updates of GD or SGD lie. They found that the path of the gradient flow %by discrete updates 
does not follow the original loss, but a \textit{modified loss} under the influence of the small yet finite learning rate as well as the optimization method itself. The approximation of the modified loss function consists of two terms, the original loss term and the implicit regularizer term proportional to the learning rate. The implicit regularizer leads GD and SGD to flat minima by penalizing the norm of the gradient, thus contributing to the generalization.

% We tried to compute the implicit regularizer for FedAvg on non-IID data. 
Unlike GD and SGD, we found that its implicit regularizer works negatively for the convergence of FedAvg. If we assume that the number of local SGD steps ($E$) and the learning rate ($\eta$) are finite but small enough to disregard the high-order terms $O(E^3 \eta^3)$ in the parameter updates, we found that the implicit regularizer term of FedAvg has a different form from SGD's. That is, the first-order term of the implicit regularizer contains a special term, which we call the \textit{dispersion term}. 
When the modified loss is minimized, the dispersion term is also minimized, which increases the distance of each client's gradient from the average gradient. Independently from this variance, 
we found through experiments that the dispersion term affects the convergence of FedAvg. 
%analyzed this bias caused by the dispersion term and showed that such a bias can hamper the convergence of FedAvg. 
We also found that there is another term in the higher-order terms of the implicit regularizer (we call the \textit{secondary dispersion term}), which can affect the convergence of FedAvg as well. 
So, both terms make the parameter updates deviate from the original path, which we call a \textit{bias} in this paper. 
% 해당 문구에 대해 김수현 박사님과 상의 필요

We also obtained the implicit regularizer for variance reduction methods such as \textit{FedSAM} \citep{caldarola2022fedasam, qu2022generalized} and \textit{SCAFFOLD} \citep{karimireddy2020scaffold}. We found that they mitigate the problem of the dispersion term by enforcing the gradients to be close to each other, thus automatically reducing the drifts. 
%Fortunately, many variance reduction methods like FedSAM \citep{caldarola2022fedasam, qu2022generalized} and SCAFFOLD \citep{karimireddy2020scaffold} can mitigate the dispersion term problem: they enforce the gradients to be close to each other and automatically reduce the drifts. 
Their convergence analysis based on convex optimization clearly showed their advantage.
%heir Those works have done convergence analyses to show the advantages of their algorithms.
However, our analysis based on the implicit regularizer of FedSAM and SCAFFOLD can show their limitations as well as their strengths.
We can prove that FedSAM can partially remove the bias %effect 
for the dispersion term. % and cannot perform well as SGD. 
On the other hand, SCAFFOLD can completely remove the bias for the dispersion term, but not for the second-order terms of the implicit regularizer.
Generally, the implicit regularizer-based analysis is useful since it can provide an insight on a more complex, non-convex setting where federated learning is often employed. Actually, there is an analysis of SCAFFOLD in a non-convex setting, which gave a good mathematical insight on its convergence speed \citep{karimireddy2020scaffold}. Compared to that, our analysis can provide a more intuitive insight on where it converges and why. For example, the implicit regularizer of FedAvg can also explain sharpness of the loss surface at the sub-optimal point where it converges.

%ts limitation emerges in the second-order terms in its implicit regularizer. 

\paragraph{Contributions.} Our main contributions are as follows.
\begin{itemize}
    \item This is the first work to analyze the implicit regularizer for federated learning on non-IID data, which can explain convergence on complex, non-convex settings. 
    %which can explain the bias of FedAvg more in-depth than existing convergence analyses.
    \item The implicit regularizer of FedAvg not only can explain its defect in convergence but also the sharpness of the loss surface where FedAvg converges.
    %The implicit regularizer of FedAvg not only can explain its defect in convergence but where FedAvg converges.
    %, to provide both empirical and theoretical explanation of how implicit regularization of FedAvg affects the convergence behaviour. 
    % \item Through a backward error analysis, we analyze the implicit regularizer of FedAvg to provide both empirical and theoretical explanation of how implicit regularization of FedAvg affects the convergence behaviour.
    \item The implicit regularizer of FedSAM and SCAFFOLD can explain their limitations as well as their advantages.
    % \item Future FL algorithms can benefit from our work by examining the dispersion term from their implicit regulizer. 
    %\item We provided thorough empirical analysis as well as rigorous theoretical one to prove above
\end{itemize}

\section{Related Work}
\label{Related work}
\paragraph{Convergence of FedAvg on non-IID data.}

It is widely known that FedAvg converges slower than centralized learning. Previous work such as \citet{stich2018local}, \citet{patel2019communication}, and \citet{khaled2020tighter} have analyzed the convergence behaviour of FedAvg under a different name, Local SGD. Such a slow convergence becomes more severe when the data distribution of the clients is non-IID \cite{karimireddy2020scaffold}.  
%heterogeneous and the analysis needs to be more delicate \cite{karimireddy2020scaffold}. 
Many researches such as \citet{zhao2018noniid}, \citet{yu2019parallel}, \citet{wang2019adaptive}, \citet{haddadpour2019convergence}, and \citet{li2020convergence} have done a sharp analysis on such a situation, focusing on the asymptotic convergence speed of FedAvg in the best and worst cases. In addition to the slow convergence, \citet{li2020convergence} focused on the inherent bias of FedAvg.
They showed that the path of parameter updates in FedAvg deviates from the path of SGD, deriving a model whose performance is lower than the SGD's.
%the converged model might be subpar than the converged model from SGD.

\paragraph{Implicit regularization.}

Implicit regularization \citep{barrett2021implicit} \citep{smith2021on} is a key concept used in this paper. Using backward error analysis, one can find the path the gradient flow actually takes under the influence of implicit regularization, defined by a finite learning rate and an optimization method itself such as GD or SGD.
In this work, we extend implicit regularization to the federated learning domain to analyze the gradient flow of FedAvg. 

%While the concept was originally used to assess the modified flow of GD and SGD, we adapt it to analyze the gradient flow of FedAvg. 

As a similar approach to our analysis, \citet{glasgow2022sharp} and \citet{gu2023why} employ Stochastic Differential Equation-based approximation to analyze the gradient flow of FedAvg (Local SGD). \citet{gu2023why} observes the long-term behavior of local minimizers in FedAvg and concludes that the local minimizer of FedAvg is biased towards flat minima for IID data. Unlike this work, we focus on the short-term behavior of FedAvg on non-IID data.

\section{Backward Error Analysis}
\label{Backward error analysis}

\subsection{The idea of backward error analysis}
\label{The idea of backward error analysis}

This whole subsection briefly explain the idea of backward error analysis done in \citet{barrett2021implicit} and \citet{smith2021on}. 
The basic idea starts by considering GD as the integration of an ODE in the form $\dot{\omega} = -\nabla \loss(\omega)$, called the \textit{gradient flow}. Then discrete updates of GD can be seen as solving the integration problem with explicit Euler method of the first order, as $\omega(t+\eta) \approx \omega(t) - \eta \nabla \loss(\omega(t))$. When the step size is finite, there will be a gap between a discrete solution of a GD step and the exact solution of the gradient flow equation. 
%discreteness of GD steps will bring about discrepancy from the exact solution of differential equation. 
To bridge the gap, we introduce a modified flow of $\dot{\omega} = \tilde{f}(\omega)$ that the optimization such as GD really follows, where $\tilde{f}(\omega)$ is expressed in powers of step size $\eta$.
\begin{equation}
\label{eqf}
    \tilde{f}(\omega) = f(\omega) + \eta f_1(\omega) + \eta^2f_2(\omega) + \dots
\end{equation}
This is viable when the step size $\eta$ is finite but relatively small. Now, the role of backward error analysis is to find the function for each correction term $f_i(\omega)$. In \citet{barrett2021implicit}, the second-derivative of the parameter is
\begin{equation}
    \ddot{\omega}(t) = \nabla \tilde{f}(\omega(t)) \dot{\omega}(t) = \nabla \tilde{f}(\omega(t)) \tilde{f}(\omega(t)),
\end{equation}
and by using this, we can obtain the perturbed parameter $\omega(t + \eta)$ taking the Taylor expansion of $\tilde{f}$.
\begin{align}
    &\omega(t + \eta) \nonumber \\
    =& \omega(t) + \eta \tilde{f}(\omega(t)) + \frac{\eta^2}{2} \nabla \tilde{f}(\omega(t)) \tilde{f}(\omega(t)) + O(\eta^3)
    \label{eqwt}
\end{align}
From Equation \ref{eqf}, it is possible to express $\tilde{f}(\omega(t))$ with the original function $f(\omega(t))$ and correction terms $f_1(\omega(t))$, $f_2(\omega(t))$, $\dots$, and modify Equation \ref{eqwt} as
\begin{align}
\textstyle
    \omega(t+\eta) &= \omega(t) + \eta f(\omega(t)) + \eta^2 (f_1(\omega(t)) \nonumber \\
    &+ \frac{1}{2} \nabla f(\omega(t)) f(\omega(t))) + O(\eta^3).
    \label{eqwte}
\end{align}
In order to derive the correction terms, we need to ensure the equivalence between the parameter from the continuous modified flow and the parameter discretely updated during a single or multiple steps of optimization. 
% In order to derive the correction terms, the parameter obtained from the continuous modified flow should be the same as the parameter discretely updated during a single or multiple steps of optimization.
For GD, as a simple example, the parameter after a single discrete update, $\omega_{t+1} = \omega_t - \eta \nabla \loss(\omega_t)$, should be the same as  the parameter, $\omega(t+\eta)$. Then $f(\omega)$ can be fixed as $-\nabla \loss(\omega)$. Also, at order $\eta^2$, the coefficient must go to zero which means that
\begin{equation}
    f_1(\omega) = -\frac{1}{2} \nabla \nabla \loss(\omega) \nabla \loss(\omega) = -\frac{1}{4} \nabla \|\nabla \loss(\omega)\|^2.
\end{equation}
Finally, it is possible to say that discrete GD iterates following the path of an ODE with the form of
\begin{equation}
    \dot{\omega} = -\nabla \loss (\omega) - (\eta/4) \nabla \|\nabla \loss(\omega)\|^2 + O(\eta^2).
\end{equation}
If the step size $\eta$ is small enough to ignore the high order terms, the modified loss that the parameter updates of GD truly follows can be expressed as
\begin{equation}
    \tilde{\loss}_{GD}(\omega) \approx \loss (\omega) + \frac{\eta}{4} \|\nabla \loss(\omega)\|^2.
\end{equation}
The modified loss is regarded as the the true loss function that the parameter updates of GD should minimize.

The same technique can be applied to SGD. In this case, rather than  one step, a single epoch is taken into consideration to obtain the modified loss where $\nabla \loss_k (\omega)$ is the loss of $k$-th mini-batch in $E$ iterations of one epoch:
\begin{equation}
    \tilde{\loss}_{SGD}(\omega) \approx \loss (\omega) + \frac{\eta}{4E} \sum^{E-1}_{k=0} \|\nabla \loss_k(\omega)\|^2.
\end{equation}

\subsection{Backward error analysis for FedAvg}
\label{Backward error analysis for FedAvg}

%The same backward error analysis technique can be used to explain the convergence behavior of FedAvg. Unlike in central learning, in FedAvg, it is the client that trains the model with training samples. With their own data, the clients locally update their parameters for multiple steps and send their parameters to the server. What the server does is aggregating those local parameters to produce a new global parameter. In FedAvg, aggregation is done simply by averaging those local parameters. A single iteration of training and aggregation is called a \textit{round}. Therefore, unlike in GD where only a single step is taken into consideration, we take the result of multiple steps in one round to match the solution on the modified continuous flow. 

The same backward error analysis can be used to explain the convergence behavior of FedAvg. Unlike in central learning, in FedAvg, it is the client that trains the model with training samples. With their own data, the clients locally update their parameters with multiple steps (one step means one iteration of parameter update) and send their updated parameters to the server. What the server does is aggregating (averaging in FedAvg) those local parameters to produce new global parameters. %In FedAvg, aggregation is done simply by averaging those local parameters. 
A single iteration of training and aggregation is called a \textit{round}. Therefore, unlike in GD where we try to match the solution of a single step, we try to match the solution of the multiple steps of one round to the modified continuous flow. Pseudocode for FedAvg is in Appendix.

We first define the necessary variables below, assuming that FedAvg runs with full participation of the clients.

\paragraph{Notations.}
The number of clients is $m$ and the number of local steps in one round is $E$. The parameter is $\omega$ and local parameters are updated with a finite learning rate $\eta$.
The loss function of the mini-batch sample from $k$-th step of $j$-th client is defined as $\loss_{jk}(\omega)$. The mean loss function of each client is defined as $\loss_{j}(\omega) = \frac{1}{E} \sum_{k=0}^{E-1} \loss_{jk}(\omega)$, and $\loss(\omega) = \frac{1}{m} \sum_{j=0}^{m-1} \loss_{j}(\omega)$. $\nabla \loss_{j}(\omega)$ is called a \textit{client gradient} and $\nabla \loss(\omega)$ is called the \textit{global gradient} in this paper.

Now we assume that the learning rate $\eta$ is large enough to make $O(E^2 \eta^2)$ significant, yet too small to make $O(E^3 \eta^3)$ significant.
With these assumptions and definitions, we obtain the loss function modified under FedAvg as follows.

\begin{theorem}
    If local parameters of clients are discretely updated with a finite learning rate, the expectation of discrete updates of the aggregated parameter in FedAvg follows the modified loss \(\tilde{\loss}_{FedAvg} (\omega)\) which can be expressed as
    \begin{align}
        \tilde{\loss}_{FedAvg}(\omega) &\approx \loss(\omega) - \frac{E\eta}{4m} \sum_{j=0}^{m-1} \underbrace{\|\nabla \loss(\omega) - \nabla \loss_{j}(\omega)\|^2}_{\text{Dispersion term}} \nonumber \\
        &+ \frac{\eta}{4mE} \sum_{j=0}^{m-1} \sum_{k=0}^{E-1} \|\nabla \loss_{jk}(\omega)\|^2.
        \label{eq:first-order}
    \end{align}
    The approximation holds when $\eta \ll 1 / E$. If $E=1$, the modified loss is the same as the one of SGD.
\end{theorem}

\paragraph{Dispersion term.}

Unlike SGD, the implicit regularizer of FedAvg is composed of two terms. The latter term, which is the same as in the implicit regularizer of SGD, is the one known to aid generalization and help the converged model to achieve a high accuracy \citep{smith2021on}. One thing to note is that the latter term is affected by the size of each mini-batch, rather than the effective batch size \cite{lin2019don} which is a size of a mini-batch multiplied by the number of local steps $E$. The former term, called \textit{dispersion term} in this paper, is the one that makes a difference. Equation \ref{eq:first-order} indicates that when the modified loss is reduced (so is the the dispersion term), the dispersion term increases the distance between the client gradient and the global gradient. 
%From the dispersion term, an undesirable bias can be induced and 
%the presence of 
%such a bias can affect the convergence seriously, which we will show through experiments.
% Independently,
The presence of the dispersion term can affect convergence severely, which we will show through experiments.

% orthogonal bias에 대한 얘기 추가

\paragraph{Dispersion term and sharp minima.}

Many optimization problems such as minimization of cross-entropy can be regarded as a Maximum Likelihood Estimation problem. Such a problem setting makes an interesting point about the Hessian of the loss. If we assume that the current parameter $\omega$ is close to an optimum and the outputs of a model in the current parameter are almost identical to the ground-truth, the Hessian of the loss can be approximated as Fisher information matrix. Moreover, since the loss gradient is almost zero nearby optima, we can build a special equivalence of the trace of Fisher information matrix and the gradient variance of samples. %From the fact that 
Since we can approximate the Hessian of loss as Fisher information matrix, we can deduce that the trace of Hessian can be approximated by the gradient variance of samples \cite{rame2022fishr}. If the number of samples in the dataset is $N$ and the loss gradient of the $i$-th sample is $\nabla \tilde{\loss}_i(\omega)$, the trace of Hessian is approximated as 
\begin{equation}
    \textstyle
    \mathrm{tr}(\nabla \nabla \loss(\omega)) \approx \frac{1}{N} \sum^N_{i=1} \| \nabla \tilde{\loss}_i(\omega) - \nabla \loss(\omega) \|^2
    \label{eqtrace}
\end{equation}
Meanwhile, if one considers that the dispersion term of FedAvg on non-IID data increases the gradient variance, it is possible to build another link between FedAvg and the Hessian trace. When the parameter is nearby optima, the dispersion term of FedAvg can increase the trace of Hessian. Since the trace of Hessian can be a measure for \textit{sharpness} of the loss surface \cite{ma2021linear}, it can be stated that the dispersion term leads the parameter to sharp minima
(later, we will see an opposite characteristics of FedAvg, though).
% 해당 문구에 대해 김수현 박사님과 상의 필요
% Though convergence into sharp minima does not necessarily mean worse convergence \cite{dinh2017sharp}, a sharp minimum that FedAvg falls into is a 

\subsection{Backward error analysis for FedSAM}

% sharp minima에 대한 얘기 추가

While the dispersion term of FedAvg on non-IID data drives the parameter to converge into sharper minima, there is a well-known technique called \textit{sharpness-aware minimization} (SAM) that can produce an opposite effect. SAM is a technique that alters the loss function into a form of
\begin{equation}
    \loss_{SAM} (\omega) = \loss(\omega + \varepsilon \nabla \loss (\omega)) \text{ where } \varepsilon \ll 1
\end{equation}
% Sharpness-aware minimization (SAM) is a good tool to make such an effect and is able to lead the parameter to a flatter minimum. By guiding the parameter to a flatter minimum, SAM can mitigate the bias of the dispersion term towards sharp minima and 
and it is already well-known that SAM increases the convergence speed of FedAvg \citep{caldarola2022fedasam, qu2022generalized}. 
% Such a result contrasts with the behaviour of secondary dispersion terms that also guide the parameter to flat minima but reduce the convergence speed of FedAvg. 
However, we analyze the implicit regularization of FedSAM to show that SAM acts as a variance reduction method by partially removing the dispersion term, rather than na\"ively leading the parameter to flatter minima.

% Though seeking flat minima seems far from removal of the dispersion term, we suggest that seeking flat minima is highly related to mitigating the dispersion effect. In this section, we analyze how FedSAM \citep{caldarola2022fedasam, qu2022generalized} acquires better performance than FedAvg.

For clarity, we refer to an approximation method in \citet{zhao2022penalizing} and \citet{geiping2022stochastic} that is used for computation of Hessian-vector product. Through a finite-difference approximation, the corresponding gradient to a gradient norm penalty is computed as
$\nabla \frac{1}{2} \|\nabla \loss(\omega)\|^2 \approx \frac{\nabla \loss(\omega + \varepsilon \nabla \loss(\omega)) - \nabla \loss(\omega)}{\varepsilon}$
when the magnitude of $\varepsilon$ is very small like $0.01 / \|\nabla \loss(\omega)\|$ as in \citet{foret2021sharpnessaware}. If we put this to a loss as an explicit regularizer with coefficient $\lambda$,
\begin{align}
    &\textstyle \loss(\omega) + \frac{\lambda}{2} \|\nabla \loss(\omega)\|^2  \nonumber \\
    \approx&\textstyle \frac{\lambda}{\varepsilon} \loss(\omega + \varepsilon \nabla \loss(\omega)) + (1 - \frac{\lambda}{\varepsilon}) \loss(\omega)
\end{align}
If $\lambda = \varepsilon$, the cost function is the same as the one of SAM. This implies that SAM is equivalent to training a model with a gradient norm penalty. However, $\varepsilon$ of SAM could be smaller than $E \eta / 2$ if the norm of the gradient is large. Also, FedSAM gives penalty to the mini-batch gradients instead of the norm of an average gradient of mini-batches of a client. 
In this case, if a subsidiary implicit regularization caused by gradient penalty itself can be ignored, the result below holds.
\vspace{-0.5em}
\begin{corollary}
    If local parameters are updated with a finite learning rate, the expected path of the global parameter updates of FedSAM approximately follows the modified loss \(\tilde{\loss}_{FedSAM} (\omega)\) as
    % \vspace{-1em}
    \begin{align}
    \textstyle
        \tilde{\loss}_{FedSAM}(\omega) &\approx \loss(\omega) + \frac{\varepsilon}{2} \|\nabla \loss(\omega)\|^2 \nonumber \\
        &- (\frac{E\eta}{4m} - \frac{\varepsilon}{2m}) \sum_{j=0}^{m-1} \|\nabla \loss(\omega) - \nabla \loss_{j}(\omega)\|^2 \nonumber \\
        \textstyle
    & + \frac{\varepsilon}{2mE} \sum_{j=0}^{m-1} \sum_{k=0}^{E-1} \|\nabla \loss_{j}(\omega) - \nabla \loss_{jk}(\omega)\|^2
    \label{eqsam}
    \end{align}
    \label{theorem:fedsam}
    The approximation holds when $\eta \ll 1 / E$.
\end{corollary}

The modified loss of FedSAM is under the influence of $\varepsilon$. It can be observed that the presence of $\varepsilon$ reduces the magnitude of the dispersion term. This indicates that FedSAM can function as a variance reduction method that decreases the dispersion of gradients. 
% Furthermore, the third term of the implicit regularizer diminishes the batch gradient variance, which can in turn enhance the convergence speed. 
It is important to note, however, that if $\varepsilon$ is smaller than $E \eta / 2$, such effects by $\varepsilon$ will only \textit{partially} remove the dispersion term and might not sufficiently enhance the performance of the model as other variance reduction methods. We will evaluate it through experiments.
% We verify this through several experiments and the results are in Appendix \ref{experiment:fedsam}.

% We empirically check that the such a bias can hamper the convergence of FedAvg in Section \ref{experiment:first}.

\subsection{Backward error analysis for SCAFFOLD}

Another simple solution to mitigate the effect of the dispersion term would be enforcing the gradients to be close to each other. A variance reduction method such as SCAFFOLD \cite{karimireddy2020scaffold} is the one. It utilizes the control variates to remove the dispersion effect of FedAvg. The convergence behavior of SCAFFOLD is quite different from FedAvg's, and as done for FedAvg, a backward error analysis can be done for SCAFFOLD to explain its convergence behavior. For convenience, we again assume full participation of all clients. If SCAFFOLD is under an ideal situation and the learning rate is small enough to make $O(E^2 \eta^2)$ negligible, we obtain the modified loss of SCAFFOLD.

\begin{theorem}
    If local parameters of clients are discretely updated with a finite learning rate, the expectation of discrete updates of the aggregated parameter in SCAFFOLD follows the modified loss \(\tilde{\loss}_{SCAFFOLD} (\omega)\) which is expressed as
    \begin{align}
    &\tilde{\loss}_{SCAFFOLD}(\omega) \approx \loss(\omega) + \frac{\eta}{4} \|\nabla \loss(\omega)\|^2 \nonumber \\
    & + \frac{\eta}{4mE} \sum^{m-1}_{j=0} \sum^{E-1}_{k=0} \|\nabla \loss_j(\omega) - \nabla \loss_{jk}(\omega))\|^2
    \label{eq:scaffold}
    \end{align}
    The approximation holds when $\eta \ll 1 / E$.
\end{theorem}

The implicit regularizer of SCAFFOLD consists of two terms, $\frac{\eta}{4} \|\nabla \loss(\omega)\|^2$ and $\frac{\eta}{4mE} \sum^{m-1}_{j=0} \sum^{E-1}_{k=0} \|\nabla \loss_j(\omega) - \nabla \loss_{jk}(\omega))\|^2$. The former term has the same form as the implicit regularizer of GD and penalizes the norm of the global gradient. The latter penalizes the trace of covariance matrix of local mini-batch gradients from each client, unlike the implicit regularizer of SGD that penalizes the trace of covariance matrix of all mini-batch gradients \cite{smith2021on}. 

The most notable point is that the dispersion term is \textit{absent} in the modified loss of SCAFFOLD. SCAFFOLD is mostly not affected by the effect of the dispersion term and its convergence behaviour resembles SGD more than FedAvg, which can be verified in the experiments.
% Still, the control variates that remove the dispersion term are approximations of the gradients from the previous round, and SCAFFOLD can be affected by high-order terms of the modified loss.

\subsection{Empirical analysis}

One way to inspect the effect of the dispersion term is to compare the convergence behaviour of FedAvg with and without the dispersion term. In order to empirically check the effect of the dispersion term, we manually removed the dispersion term from the modified loss. 
% We first calculated the average of client gradients $\nabla \loss(\omega_0)$ and $\nabla \|\nabla \loss_i(\omega_0)\|^2$ in the server and sent them to the clients. In the clients, we calculated the average of its mini-batch gradients $\nabla \loss_{j}(\omega_0)$. We then obtained $\frac{E^2 \eta^2}{4} (\nabla \| \nabla \loss_{j}(\omega_0)\|^2 - \nabla \|\nabla \loss(\omega_0)\|^2)$ and subtracted it from the parameter of the client. The modified loss of the aggregated parameter now is $\tilde{\loss}_{modified}(\omega) = \loss(\omega) + \frac{\eta}{4mE} \sum_{j=0}^{m-1} \sum_{k=0}^{E-1} \|\nabla \loss_{jk}(\omega)\|^2$.
The algorithm for removing the dispersion term is in the Appendix.

\begin{figure*}[]
    \centering
    \subfigure[]{
        \includegraphics[width=.24\textwidth]{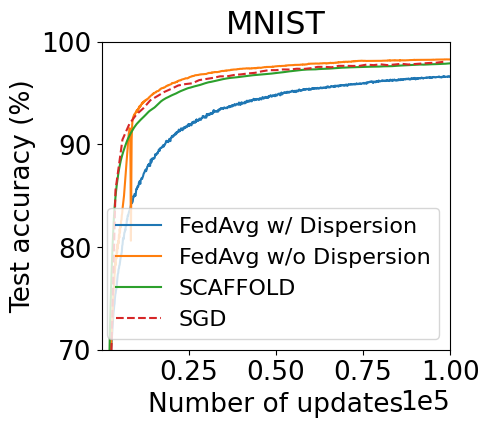}
    }
    \hfill
    \subfigure[]{
        \includegraphics[width=.24\textwidth]{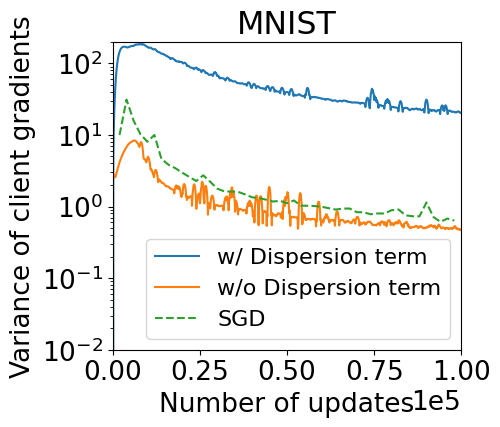}
    }
    \hfill
    \subfigure[]{
        \includegraphics[width=.23\textwidth]{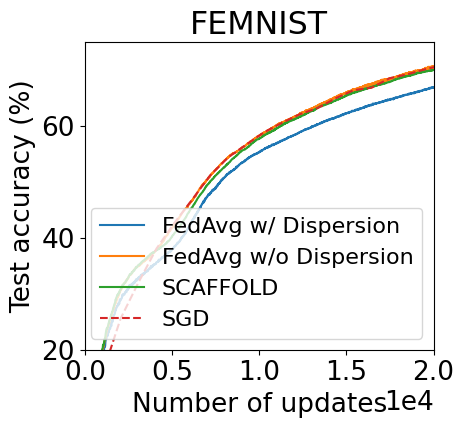}
    }
    \hfill
    \subfigure[]{
        \includegraphics[width=.235\textwidth]{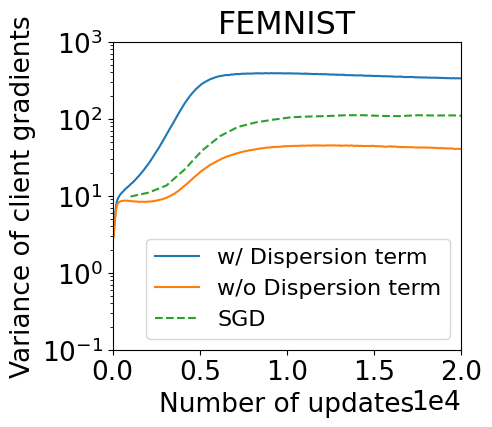}
    }
    \caption{Test accuracy and variance of client gradients of FedAvg, SCAFFOLD, and SGD on MNIST, and FEMNIST. The final test accuracy is higher and the variance of client gradients is significantly lower when the dispersion term is absent in the modified loss. The convergence behaviours of SCAFFOLD, SGD, and FedAvg without dispersion term are almost identical.}
    \label{FigDrift}
\end{figure*}

\paragraph{Empirical analysis on the dispersion term.}
\label{experiment:first}

We run experiments for evaluation of our analysis. To evaluate only the effect of the dispersion term, we run experiments with a simple CNN model on a simple dataset, MNIST \citep{lecun1998gradient} and a relatively more complex dataset, FEMNIST \citep{caldas2018leaf}. Experiments were done on a non-IID environment of Dirichlet distribution with parameter 0.2, except for FEMNIST, which is naturally non-IID. The batch size was 30 for MNIST and 100 for FEMNIST. The effective batch size mentioned in \textbf{Dispersion term.} was not applied to SGD and the batch size was set the same for both FedAvg and SGD. This is to make the implicit regularizer of FedAvg the same as the one of SGD other than the dispersion term. To fully observe the effect under the gradient flow, we used a normal SGD optimizer with a small learning rate of 0.001 with no momentum and learning rate decay. More details on the experimental settings such as the model architecture and the learning rate are in the Appendix.

The empirical results are summarized in Figure \ref{FigDrift}. Overall, FedAvg without the dispersion term converged faster than the original FedAvg and the performance was identical to the ones of SGD and SCAFFOLD, whose modified loss is similar to the one of FedAvg without the dispersion term. This indicates that the dispersion term is the main reason for performance degradation of FedAvg when the size of a learning rate is small enough. Also, the variance of the client gradients, measured at the beginning of the round, was much higher when there was a dispersion term in the modified loss, which matches our theoretical observation. These result indicates that our analysis on the first-order term of implicit regularizer can explain the convergence behaviour of FedAvg and SCAFFOLD. However, one thing to note is that the main reason for performance degradation of FedAvg is \textit{not the noise of gradients} due to the increased gradient variance, but rather the presence of %bias due to the 
the implicit regularizer itself. We will discuss this in later sections.

\begin{figure}
    \centering
    \subfigure[]{
        \includegraphics[width=.21\textwidth]{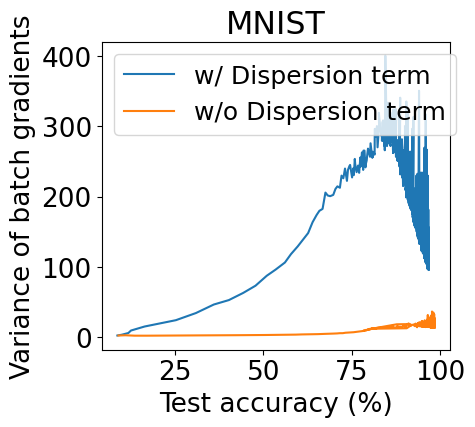}
    }
    \subfigure[]{
        \includegraphics[width=.21\textwidth]{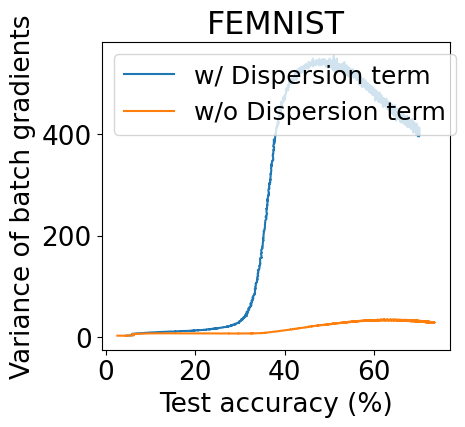}
    }
    \caption{Variance of mini-batch gradients in MNIST and FEMNIST.}
    \label{fig:femnistbatch}
\end{figure}

\paragraph{Implicit bias of FedAvg.}

In Figure \ref{FigDrift} and \ref{fig:femnistbatch}, FedAvg with the dispersion term has higher variances of mini-batch gradients as well as client gradients than  FedAvg without it, even when the performance of the model is the same. If the updates of FedAvg with the dispersion term slowly but surely followed the same path as FedAvg without the dispersion term, the variance of gradients should have been the same when the performance of the model was the same. However, the difference in variance implies that the parameter updates of FedAvg deviated from the original path due to the bias by implicit regularization. Such a deviation heavily affects the convergence behavior and leads FedAvg to converge into a sub-optimal solution as pointed out by \citet{li2020convergence}. 
% Also, an increased variance of mini-batch gradients due to the dispersion term makes the optimizer to flutter and forces FedAvg to become slower than other optimization methods.

% Another thing to notice is that the gradient variance of SGD was consistently larger than that of FedAvg without the dispersion term, even when the test accuracy of the model was the same. This indicates that high-order terms in the implicit regularizer of FedAvg, such as secondary dispersion terms, reduce the variance of gradients. Further results will be inspected in-depth in the next section.

\paragraph{Empirical analysis on FedSAM.}
\label{experiment:fedsam}

We experimented with FedSAM on MNIST and Fashion-MNIST \cite{xiao2017fashion} on non-IID with full client participation. We used simple datasets to observe the effect of the dispersion term without a disturbance of batch gradient variance. More details on the experimental settings are in the Appendix.

\begin{figure}[t]
    \centering
    \subfigure[]{
        \label{fig:vareps_mnist}
        \includegraphics[width=.2\textwidth]{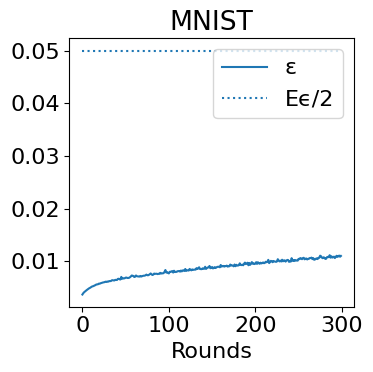}
    }
    \subfigure[]{
        \label{fig:vareps_fmnist}
        \includegraphics[width=.2\textwidth]{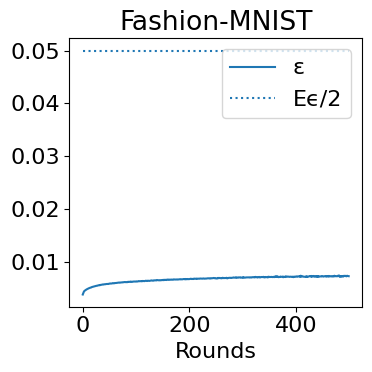}
    }
    \caption{The value of $\varepsilon$ of FedSAM on MNIST and Fashion-MNIST. $\varepsilon$ is consistently lower than $E\eta / 2$.}
    \label{FigSAMvareps}
    % 공지. epsilon to eta
\end{figure}

\begin{figure*}[t]
    \centering
    \subfigure[]{
        \label{fig:acc_mnist}
        \includegraphics[width=.19\textwidth]{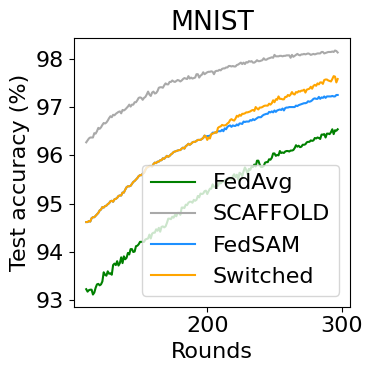}
    }
    \hfill
    \subfigure[]{
        \label{fig:var_mnist}
        \includegraphics[width=.19\textwidth]{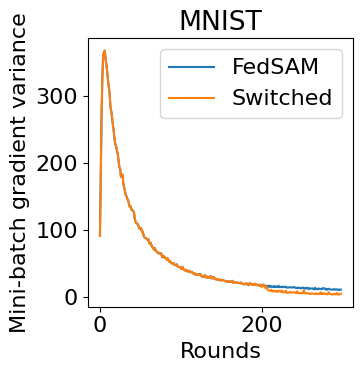}
    }
    \hfill
    \subfigure[]{
    \label{fig:acc_fmnist}
        \includegraphics[width=.19\textwidth]{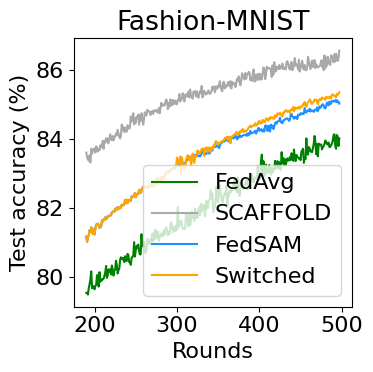}
    }
    \hfill
    \subfigure[]{
        \label{fig:var_fmnist}
        \includegraphics[width=.19\textwidth]{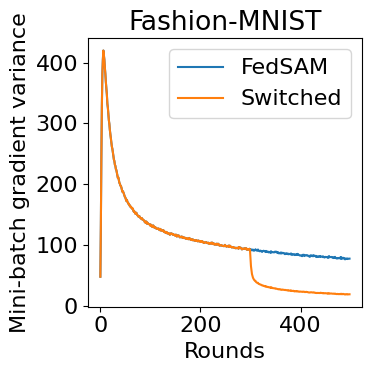}
    }
    \caption{Test accuracy, value of $\varepsilon$, and client gradient variance of FedSAM on MNIST and Fashion-MNIST. $\varepsilon$ was switched to $E \eta / 2$ during training and the convergence speed became faster while the variance of mini-batch gradients decreased. The values at the exact round where switching occurred were omitted for smoothing of the graphs.}
    \label{FigSAM}
    % 공지. epsilon to eta
\end{figure*}

We first investigated the value of $\varepsilon$ during a normal training with FedSAM. The value of $\varepsilon$ was set as the value similar to $0.01 / \sqrt{\|\nabla \loss_{jk} (\omega)\|}$, which was much more stable than $0.01 / \|\nabla \loss_{jk} (\omega)\|$. A detailed explanation is in the pseudocode in the Appendix. As shown in Figure \ref{fig:vareps_mnist} and \ref{fig:vareps_fmnist}, the value of $\varepsilon$ stayed below $E \eta / 2$. 

Next, we examined the performance of FedSAM. We checked if FedSAM could perform as well as SCAFFOLD even when the value of $\varepsilon$ stays below $E \eta / 2$. As shown in Figure \ref{fig:acc_mnist} and \ref{fig:acc_fmnist}, FedSAM was able to perform better than FedAvg but not as well as SCAFFOLD. This accords with our prediction that FedSAM would only partially mitigate the dispersion effect if $\varepsilon$ is not large enough, while SCAFFOLD is able to almost remove the dispersion term.

To confirm if the slow convergence of FedSAM is due to the insufficient magnitude of $\varepsilon$, we changed the value of $\varepsilon$ in Equation \ref{eqsam} to $E \eta / 2$ during training, which makes the magnitude of the dispersion term zero. We increased the value of $\varepsilon$ to $E \eta / 2$ in the middle of training session when the gradients are relatively stable. We did not change $\varepsilon$ in the early stage of training since the gradient variance tends to increase rapidly in the early stage and the gradients become extremely unstable. We switched $\varepsilon$ at the 200-th round on MNIST and at the 300-th round on Fashion-MNIST.

As shown in Figure \ref{fig:acc_mnist}, \ref{fig:acc_fmnist}, the performance of FedSAM with switched $\varepsilon$ surpassed the performance of the original FedSAM due to a faster convergence speed. The result indicates that the convergence speed becomes faster when $\varepsilon$ is as big as $E \eta / 2$ and the dispersion term is fully removed. Also, as shown in \ref{fig:var_mnist} and \ref{fig:var_fmnist}, the dispersion term was mitigated, and the variance of client gradients was decreased after switching, consistent with our analysis.

% SCAFFOLD가 fully remove한단 얘기 좀 하시오

\section{High-order Terms in the Implicit Regularizer}

Meanwhile, one thing was unexplained in the empirical results on the dispersion term. In Figure \ref{FigDrift}, the variance of client gradients in SGD was higher than the one of FedAvg without the dispersion term. It means that there is still an unexplained bias of FedAvg which reduces the client gradient variance. Here, we analyze the implicit regularizer of FedAvg with \textit{high-order terms} and show that those high-order terms can reduce the client gradient variance but hamper the convergence of FedAvg. For our analysis, we use the notion of \textit{local epochs}.
To reduce the communication cost between the client and the server, federated learning often employs local epochs. That is, clients iterate over their entire local data multiple times, and one full iteration is called one \textit{local epoch}. 
Since the presence of multiple local epochs increase the number of total local steps $E$, the high-order terms $O(E^3 \eta^3)$ become significant. 

Here, the number of total local steps $E$ can be obtained with two factors: the number of local epochs and the number of local steps within a single local epoch, defined as $a$ and $K$, respectively. In this paper, we deal with a case where the number of steps per a local epoch, $K$, is small enough to ignore $O(K^3 \eta^3)$, while the number of local epochs, $a$, is large enough to make $O(a^3 K^3 \eta^3)$ significant. Now, we ignore high-order terms within a local epoch such as $O(K^3 \eta^3)$, while keeping $O(a^3 K^3 \eta^3)$ to get a more precise, modified loss with the second-order terms in the implicit regularizer.

\begin{corollary}
    If local parameters are updated with a finite learning rate for multiple local epochs, the expectation of discrete updates of the aggregated parameter in FedAvg follows the modified loss \(\tilde{\loss} (\omega)\) as
    \begin{align}
        \tilde{\loss}(\omega) &\approx \loss(\omega) - \frac{aK \eta}{4m} \sum_{j=0}^{m-1} \underbrace{\|\nabla \zeta_j (\omega - \frac{aK\eta}{3} \nabla \loss_j (\omega))\|^2}_{\text{Transformed dispersion term}} \nonumber \\
        &+ \frac{a^2 K^2 \eta^2}{6m} \sum_{j=0}^{m-1} \underbrace{\nabla \zeta_j (\omega)^\top \nabla \nabla \loss (\omega) \nabla \zeta_j (\omega)}_{\text{Secondary dispersion term}}.
        \label{eqsecondary}
    \end{align}
    \label{theorem:second}
    when $\nabla \zeta_j (\omega) = \nabla \loss (\omega) - \nabla \loss_{j}(\omega)$ and $\nabla \nabla \loss (\omega)$ denotes the Hessian of the loss. The approximation holds when $1/E^2 \ll \eta \ll 1 / E$ and $a \gg 1$.
\end{corollary}

Now, the implicit regularizer of FedAvg has been transformed. The first term of the implicit regularizer, which we call the \textit{transformed dispersion term}, is now the client gradient variance at $\omega - \frac{aK\eta}{3} \nabla \loss_j (\omega)$ instead of at $\omega$. 
Due to a transformation of the dispersion term, the modified loss no longer maximizes the gradient variance at the current parameter $\omega$. 
%Such a transformation of the dispersion term no longer makes the modified loss maximize the gradient variance at the current parameter $\omega$. 
This will reduce the effect of the original dispersion term on increasing the gradient variance.
On the other hand, we refer to the latter term of implicit regularizer as the \textit{secondary dispersion term}. One thing to note is that secondary dispersion term is a quadratic objective function. When the loss Hessian is positive-semidefinite, the quadratic term minimizes the gap between the client gradients and the global gradient. 

So, both the secondary dispersion term and the difference of the dispersion term due to transformation can reduce the variance of client gradients. Experimental results below will confirm that these high-order terms of the implicit regularizer can reduce the gradient variance. Moreover, from the link between sharpness of a loss surface and the gradient variance, it can be stated that the secondary dispersion term can lead FedAvg to flat minima. 
Such an effect by high-order terms contradicts the effect of the dispersion term. The dispersion term leads FedAvg to sharp minima when the number of local steps is small as mentioned previously, while high-order terms lead FedAvg to relatively flatter minima when the number of local steps is large.
The latter aligns with \citet{gu2023why} which proves that local steps in Local SGD lead the parameter to flat minima. 

\paragraph{Gradient variance and performance.}
However, reduction of gradient variance and convergence into flat minima do not necessarily lead to better performance, as we will see in the experiments. %be seen That is, high-order terms in implicit regularizer of FedAvg also hamper convergence.
This point differs from FedSAM which also converges to flat minima. SAM partially but explicitly removes the dispersion term, whereas high-order terms of the implicit regularizer such as the transformed dispersion term %in the high-order terms 
in Equation \ref{eqsecondary} can actually induce an additional bias to FedAvg.
%original dispersion term.
Although the high-order terms of its implicit regularizer reduce the gradient variance and lead FedAvg to flat minima, they still make  parameter updates deviate from the optimal path, thus affecting convergence.
% The effect of the transformed dispersion term works against the performance of FedAvg and conceal the generalization effect of the secondary dispersion term.
% Unlike in FedSAM which also leads FedAvg to flat minima, it is possible to observe that the secondary dispersion term of FedAvg \textit{still} deflects the path of parameter updates to a `sub-optimal' direction.
% This shows that \textit{where} the parameter converges is what matters, rather than the flatness of the loss surface itself \cite{dinh2017sharp}.
% regardless of flatness of the loss surface at the converged point.
% Though the loss surface at the converged point can be flatter due to the effect of the secondary dispersion term, the entire implicit regularizer of FedAvg \textit{still} deflects the path of parameter updates to a `sub-optimal' direction. This is the point that differs from FedSAM since FedSAM partially removes the dispersion term though it also leads FedAvg to flat minima.
On the other hand, the fact that a reduction of gradient variance can lead to degradation of performance implies that the noise of the gradients from the increased gradient variance is not a critical factor in convergence of FedAvg. This is contrary to a misconception that addition of variance in gradients is a primary reason for performance degradation of FedAvg. Though the addition of variance will certainly affect the convergence speed, the empirical results show that another factor, the bias of FedAvg, is a more critical factor. 
%Finally, the dispersion term of its implicit regularizer leads FedAvg to sharp minima as mentioned in {\bf Dispersion term and sharp minima.}, while the high-order terms lead to flat minima as mentioned above, so it is the characteristics of the sub-optimal point where FedAvg converges. This is evaluated in the Appendix.

\paragraph{Limits of variance reduction methods.}
% increase bias...보다는 다른 표현으로 대체? 마크.
%aligns the gradient of the original dispersion term to the opposite direction of client gradients. Since the local parameters are updated in the opposite direction of client gradients, such a transformation will accelerate the bias by the dispersion term.
% 해당 문구에 대해 김수현 박사님과 상의 필요
Meanwhile, the fact that high-order terms in the implicit regularizer hamper the convergence is a thought-provoking point to all variance reduction methods that use stagnant gradients as the control variates. Traditional variance reduction methods such as SCAFFOLD use the client gradients from a \textit{previous} round to correct deviation of client gradients and mitigate the effect of the dispersion term. However, the client gradient that composes the transformed dispersion term is actually from the \textit{middle} of the \textit{current} round. Such a discrepancy will inevitably cause a performance degradation, which can be a limitation of variance reduction methods.
% The performance degradation will become severe especially when the learning rate and the number of local steps are not small to make $E \eta \rightarrow 1$ or the eigenvalues of the loss Hessian are not small. 
Theoretical explanation of the limitation is in the Appendix.

\begin{figure}[]
    \centering
    \subfigure[]{
        \includegraphics[width=.21\textwidth]{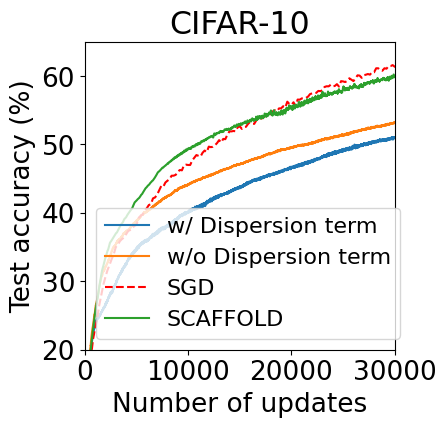}
    }
    \subfigure[]{
        \includegraphics[width=.2\textwidth]{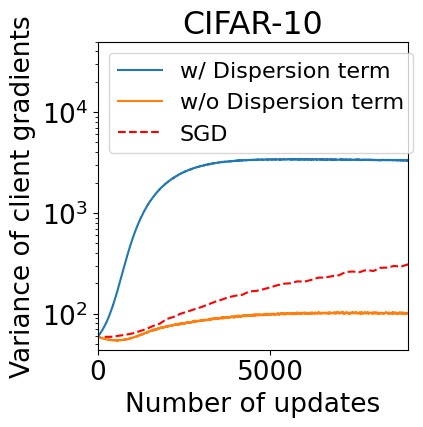}
    }
    \caption{Test accuracy and client gradient variance for a complex model and dataset. Compared to SGD that no dispersion terms, FedAvg without the first-order dispersion term has a lower accuracy but a lower variance, while SCAFFOLD has a slightly lower accuracy, due to high order terms.}
%    than SGD accuracy increases without the dispersion terms. Variance decreases without the dispersion term but increases more in SGD that has no secondary dispersion term.}
    \label{fig:cifar}
\end{figure}

\subsection{Empirical analysis on high-order terms}
\label{experiment:secondary}

We ran experiments on a rather complex dataset, CIFAR-10 \cite{krizhevsky2009learning}, for a model with residual connections to evaluate the impact of high-order terms of the implicit regularizer (small dataset and simple model do not show the impact). Data is non-IID with a Dirichlet distribution of parameter 0.05, more extreme than the previous experiment. 100 clients were trained with a learning rate of 0.001, 3 local epochs, and the batch size of 300. We experimented with FedAvg with and without the first-order dispersion term, SCAFFOLD, and SGD. Details on experimental settings are in the Appendix.

Figure \ref{fig:cifar} shows that the gradient variance of SGD, which naturally lacks the dispersion term and the secondary dispersion term in its implicit regularizer, is higher than that of FedAvg without the dispersion term. 
%is compared to the one of SGD, the variance of client gradients increased when high-order terms of implicit regularizer were naturally removed in SGD. 
This matches our analysis that the high-order terms of the implicit regularizer contributes to reducing the gradient variance.   
%in the implicit regularizer of FedAvg.
% This shows that the secondary dispersion term reduces the variance of client gradients. 
On the other hand, the convergence speed of SGD is higher although its gradient variance is higher. This indicates that the increased gradient variance itself is not the primary reason for slow convergence of FedAvg. The result also indicates that the high-order terms play a significant role in hampering convergence of FedAvg when a model is complex.
% SGD와 secondary 비교?

On the other hand, Figure \ref{fig:cifar} shows that the performance of SCAFFOLD is much higher than FedAvg's without the dispersion term and almost close to SGD's. This was expected since SCAFFOLD in Equation \ref{eq:scaffold} lacks the dispersion term of FedAvg in Equation \ref{eq:first-order}. %However, the performance of SCAFFOLD is still inferior to SGD's, which indicates the limitation of SCAFFOLD: SCAFFOLD cannot fully mitigate the bias from the high-order terms of its implicit regularizer. 
However, the performance of SCAFFOLD is still inferior to SGD's. Considering that slow-down from the batch gradient variance is larger in SGD due to more parameter updates in a single epoch, it can be concluded that SCAFFOLD cannot fully mitigate the bias from the high-order terms of its implicit regularizer.
It can be severe when a model becomes complex to make high-order terms significant, 
which has been continuously observed in \citet{reddi2020adaptive, karimireddy2020mime, yu2022tct}.

\section{Limitation and Conclusion}

% In this work, we used the backward error analysis technique to find the implicit regularizer for federated learning and analyze the convergence behaviour in a more intuitive way. The implicit regularizer of FedAvg has the dispersion term, which disperses the gradient of each client from the average gradient of clients. We showed that the bias introduced by the dispersion term deflects the path of the parameter updates from the path without the dispersion term and leads FedAvg to converge into a sub-optimal point. We also inspected into the second-order dispersion terms which hamper the convergence speed and decrease the gradient variance. Under the constraint on the accuracy of the model, the dispersion term leads FedAvg to sharp minima while the secondary dispersion terms leads FedAvg to flatter minima. The influence of the dispersion term dominates when the number of local steps are small, but the influence of the secondary dispersion terms grow as local steps increases. In contrary to a common belief, more local steps did not lead to generalization of the model.
% This suggests that more local steps might aid generalization of the model.

% However, it is true that there exists limitations in our analysis. 
As in the previous work by \citeauthor{smith2021on}, % that employs the backward error analysis technique, 
one condition required to validate our analysis is that the learning rate should be small enough to make high-order terms insignificant. Such an assumption is an extreme condition as pointed out by \citeauthor{smith2021on}. 
%the small magnitude of the learning rate. The magnitude of a learning rate should be finite but small enough to make high-order terms insignificant. 
%Such an assumption is an extreme condition as pointed out by \citeauthor{smith2021on}. 
% Especially in experiments on CIFAR-10, the maximum eigenvalue of the Hessian slightly increased as the number of local steps grows, which implies that high-order terms can have a considerable influence on the convergence behaviour.
However, a small learning rate has been commonly assumed in many federated learning researches. For example, many assume that the local learning rate is smaller than the reciprocal of the product of the number of local updates and the smoothness of the loss function \cite{karimireddy2020scaffold, qu2022generalized, xu2021fedcm}. Noting that the smoothness is the maximum eigenvalue of a Hessian, our assumption is not really extreme compared to their work. 
% In order to discard complexities, we used this assumption as a fundamental component of our analysis.

Our goal was to understand the dynamics of FedAvg and variance reduction methods in a more intuitive way than existing convex optimization-based analysis. 
Despite limitation of our assumptions, we effectively analyzed the additional implicit regularization posed %by intrinsic properties of 
federated learning methods. 
We found that the presence of the dispersion term and secondary dispersion term of the implicit regularizer of FedAvg is the main reason that affect its the convergence, rather than the noise from the increased gradient variance itself. We also analyzed the fundamental limitations of existing variance reduction methods. Empirical results confirmed our theoretical observations.
%We expect future federated learning algorithms can benefit from our work by examining the dispersion term from their implicit regularizer. 
% We were still able to investigate the strengths and limits of FedAvg and variance reduction methods, which was our main goal of this study.
% We hope our analysis aids in understanding the convergence behavior of FedAvg and variance reduction methods.
%We hope our analysis can help understandings on convergence behaviour of FedAvg that it is not the flatness of the loss surface that matters, but the bias induced by the distribution of the data environment of FedAvg.

% Employing the same backward error analysis technique, we analyzed methods that can mitigate the effect of the dispersion term: SCAFFOLD and FedSAM, and showed how effective they are when the learning rate is small enough. While SCAFFOLD was able to almost fully remove the effect of the dispersion term, FedSAM was only able to partially mitigate the dispersion effect, leaving the possibility of slower convergence.

% There exist limitations to our analysis since our analysis is valid when the learning rate is small enough to make high-order terms insignificant. Despite limitations, our analysis succeeded to provide an intuitive way to understand the behaviour of federated learning methods and the experimental results fairly corresponded to our predictions. We expect our works to be used for easy understanding of convergence behaviours under various federated learning methods.

\section{Acknowledgements}

This work was supported in part by the National Research Foundation of Korea (NRF) grant funded by Korean Government [Ministry of Science and ICT (MSIT)] under Grant RS-2023-00208245, 30\%; in part by the Institute of Information and Communications Technology Planning and Evaluation (IITP) grant funded by Korean Government (MSIT) under Grant 2021-0-00180, 20\% and Grant 2021-0-00136, 20\%; in part by the Information Technology Research Center (ITRC) support Program Supervised by IITP under Grant IITP-2021-0-01835, 20\%; and in part by IITP under Artificial Intelligence Semiconductor Support Program to Nurture the Best Talents under Grant IITP-2023-RS-2023-00256081, 10\%.

\medskip

\bibliography{aaai25}

\appendix
\onecolumn

\section{A backward error analysis on FedAvg}
\label{proof:first-order}

Firstly, we define the necessary variables beforehand. The loss function of $k$-th mini-batch sample of $j$-th client is defined as $\loss_{jk}(\omega)$. The mean loss functions are each defined as $\loss_{j}(\omega) = \frac{1}{E} \sum_{k=0}^{E-1} \loss_{jk}(\omega)$ and $\loss(\omega) = \frac{1}{m} \sum_{j=0}^{m-1} \loss_{j}(\omega)$ while the number of samples in a round and clients in a round are defined as $E$ and $m$, respectively. The parameter, $\omega^{j}$, is trained on the $j$-th client and these parameters are aggregated to form $\omega$ in the end of the round. The whole aggregated global parameter is $\omega$ and $\omega_0 = \omega(t_0)$.

For one client in one round, the learning procedure is the same as the one in SGD. Borrowing the result from \cite{smith2021on}, for $j$-th client in the round, discrete updates of the parameter $\omega^{j}$ during $E$ steps with the learning rate $\eta$ can be expressed as
\begin{align}
    \omega_E^{j} & = \omega_0^{j} - \eta \nabla \loss_{j0}(\omega_0^{j}) - \eta \nabla \loss_{j1}(\omega_1^{j}) - \eta \nabla \loss_{j2}(\omega_2^{j}) - \dots - \eta \nabla \loss_{j(E-1)}(\omega_{E-1}^{j}) \\
    & = \omega_0^{j} - \eta \sum_{k=0}^{E-1} \nabla \loss_{jk}(\omega_0^{j}) + \eta^2 \sum_{k=0}^{E-1} \sum_{l < k} \nabla \nabla \loss_{jk}(\omega_0^{j}) \nabla \loss_{jl}(\omega_0^{j}) + O(E^3 \eta^3) \\
    & = \omega_0^{j} - \eta \sum_{k=0}^{E-1} \nabla \loss_{jk}(\omega_0^{j}) + \eta^2 \xi^j(\omega_0^{j}) + O(E^3 \eta^3)
\end{align}
where $\nabla \nabla$ denotes the Hessian of a scalar function with respect to the parameter.
However, the function $\xi$ depends on the order of mini-batch samples, which makes it hard to interpret the meaning of bias. Knowing that mini-batch samples are randomly shuffled during training, instead, we obtain the expectation value of $\xi^{j}$ and get the expectation value of $\omega_E^{j}$ for easy understanding. Here we use the fact that every sequence has its own reverse order and the average of $\xi^{j}$s in a forward order and a reverse order is the same for every sequence as we can check in Equation \ref{eqorder}.
\begin{align}
    \mathbb{E}(\xi^{j}) & = \frac{1}{2} \sum_{k=0}^{E-1} \sum_{l \neq k} \nabla \nabla \loss_{jk} \nabla \loss_{jl} \\
    & = \frac{E^2}{2} \nabla \nabla \loss_{j} \nabla \loss_{j} - \frac{1}{2} \sum_{k=0}^{E-1} \nabla \nabla \loss_{jk} \nabla \loss_{jk} \\
    & = \frac{E^2}{4} \nabla(\|\nabla \loss_{j}\|^2 - \frac{1}{E^2} \sum_{k=0}^{E-1} \|\nabla \loss_{jk}\|^2) \label{eqorder}
\end{align}
\begin{align}
\label{eqa}
    \mathbb{E}(\omega_E^{j}) & = \omega_0^{j} - E \eta \nabla \loss_{j}(\omega_0^{j}) + \frac{E^2 \eta^2}{4} \nabla(\|\nabla \loss_{j}(\omega_0^{j})\|^2 - \frac{1}{E^2} \sum_{k=0}^{E-1} \|\nabla \loss_{jk}(\omega_0^{j})\|^2) + O(E^3 \eta^3)
\end{align}
The same iterations proceed on multiple clients during multiple rounds. If there are $m$ clients participating in each round, the expectation value of $\omega_E$, the parameter after one round would be
\begin{align}
\label{eqb}
    \mathbb{E}(\omega_E) &= \omega_0 - E \eta \nabla \loss(\omega_0) + \frac{E^2 \eta^2}{4m} \nabla(\sum_{j=0}^{m-1}\|\nabla \loss_{j}(\omega_0)\|^2 - \frac{1}{E^2} \sum_{j=0}^{m-1} \sum_{k=0}^{E-1} \|\nabla \loss_{jk}(\omega_0)\|^2) + O(E^3 \eta^3)
\end{align}
% Using this result, we can obtain the expectation value of $\omega_{nE}$, which is the parameter after going through $n$ rounds. But before we get $\mathbb{E}(\omega_{nE})$, we get the value of $\omega_{nE}$ itself first.
% \begin{align}
%     \omega_{nE} & = \omega_0 - E \eta \nabla \loss_0(\omega_0) + E^2 \eta^2 \xi_0 (\omega_0) - E \eta \nabla \loss_1(\omega_E) + E^2 \eta^2 \xi_1 (\omega_E) - \dots + O(E^3 \eta^3) \\
%     & = \omega_0 - E \eta \sum_{i=0}^{n-1} \nabla \loss_i(\omega_0) + E^2 \eta^2 \sum_{i=0}^{n-1} \sum_{l < i} \nabla \nabla \loss_{i}(\omega_0) \nabla \loss_{l}(\omega_0) + E^2 \eta^2 \sum_{i=0}^{n-1} \xi_i(\omega_0) + O(E^3 \eta^3) \label{eqc}
% \end{align}
% In order to obtain $\mathbb{E}(\omega_{nE})$, we combine the results from Equation \ref{eqa}, \ref{eqb}, and \ref{eqc}:
% \begin{align}
%     \mathbb{E}(\omega_{nE}) & = \omega_0 - nE \eta \nabla \loss(\omega_0) + \frac{n^2 E^2 \eta^2}{4} \nabla(\|\nabla \loss(\omega_0)\|^2 - \frac{1}{n^2} \sum_{i=0}^{n-1} \|\nabla \loss_{i}(\omega_0)\|^2 \nonumber \\ 
%     & + \frac{1}{mn^2} \sum_{i=0}^{n-1} \sum_{j=0}^{m-1}\|\nabla \loss_{j}(\omega_0)\|^2 - \frac{1}{mn^2E^2} \sum_{i=0}^{n-1} \sum_{j=0}^{m-1} \sum_{k=0}^{E-1} \|\nabla \loss_{jk}(\omega_0)\|^2) + O(n^3 E^3 \eta^3)
% \end{align}
Reminding of Equation \ref{eqwte}, since the parameter should strictly follow the modified loss $\loss_{FedAvg}$, $\mathbb{E}(\omega_{E}) = \omega(t_0 + E\eta)$ must have the form of
\begin{align}
    \mathbb{E}(\omega_{E}) &= \omega_0 - E\eta \nabla \loss(\omega_0) + E^2 \eta^2 (f_1(\omega_0) + \frac{1}{4} \nabla \|\nabla \loss(\omega_0)\|^2) + O(E^3 \eta^3)
\end{align}
Then it is possible to estimate the correction term $f_1$ to be
\begin{align}
    f_1 &= - \frac{1}{4} \nabla \|\nabla \loss\|^2 + \frac{1}{4m} \sum_{j=0}^{m-1} \nabla \|\nabla \loss_{j}\|^2 - \frac{1}{4mE^2} \sum_{j=0}^{m-1} \sum_{k=0}^{E-1} \nabla \|\nabla \loss_{jk}\|^2
\end{align}
Since an ODE has the approximated form of $\dot{\omega} = - \nabla \loss(\omega) + E \eta f_1(\omega) = - \nabla \tilde{\loss}_{FedAvg}(\omega)$, the modified loss that the gradient flow of FedAvg follows is
\begin{align}
    \tilde{\loss}_{FedAvg}(\omega) &\approx \loss(\omega) - \frac{E\eta}{4m} \sum_{j=0}^{m-1} \|\nabla \loss(\omega) - \nabla \loss_{j}(\omega)\|^2 + \frac{\eta}{4mE} \sum_{j=0}^{m-1} \sum_{k=0}^{E-1} \|\nabla \loss_{jk}(\omega)\|^2 \label{finaleq}
\end{align}
and we deduce Equation \ref{finaleq} noting that $\sum_{j=0}^{m-1} (\nabla \loss(\omega) - \nabla \loss_{j}(\omega)) = 0$. Also note that equations above are viable only when $E$ and $\eta$ are small enough to neglect high order terms: $\eta \ll 1 / E$.

For partial participation of clients where the average loss of clients in the $i$-th round is $\loss_i(\omega) = \sum^{m-1}_{j=0} \loss_{ij}(\omega)$ and the mean loss is $\loss(\omega) = \sum^{n-1}_{i=0} \loss_{i}(\omega)$, then the modified loss is
\begin{equation}
    \tilde{\loss}_{FedAvg}(\omega) \approx \loss(\omega) - \frac{E\epsilon}{4mn} \sum_{i=0}^{n-1} \sum_{j=0}^{m-1} \|\nabla \loss_i(\omega) - \nabla \loss_{ij}(\omega)\|^2 + \frac{\epsilon}{4mnE} \sum_{i=0}^{n-1} \sum_{j=0}^{m-1} \sum_{k=0}^{E-1} \|\nabla \loss_{ijk}(\omega)\|^2
\end{equation}
and $\eta \ll 1 / nE$.

\paragraph{Batch gradient variance.}

On the other hand, the analysis above depicts the expected path of the parameter updates, not the actual path of the updates. The stochastic property of the local updates induces discrepancies between the expected path and the actual path. First we work on the form of the actual path:
\begin{align}
    \xi^{j}(\omega^{j}) &= \sum^{E-1}_{k=0} \sum^{k-1}_{l<k} \nabla \nabla \loss_{jk} (\omega^{j}) \nabla \loss_{jl} (\omega^{j}) \\
    &= \sum^{E-1}_{k=0} \sum^{k-1}_{l<k} \nabla \nabla \loss_{jk} (\omega^{j}) \nabla (\loss_{jl} (\omega^{j}) - \loss_{j} (\omega^{j})) + \frac{E(E-1)}{2} \nabla \nabla \loss_{j} (\omega^{j}) \nabla \loss_{j} (\omega^{j}), \\
\end{align}
and we subtract the expected path from the actual path to get the discrepancy.
\begin{align}
    \xi^{j}(\omega^{j}) - \mathbb{E}(\xi^{j}(\omega^{j})) &= \sum^{E-1}_{k=0} \sum^{k-1}_{l<k} \nabla \nabla \loss_{jk} (\omega^{j}) \nabla (\loss_{jl} (\omega^{j}) - \loss_{j} (\omega^{j})) + \frac{E(E-1)}{2} \nabla \nabla \loss_{j} (\omega^{j}) \nabla \loss_{j} (\omega^{j}) \nonumber \\
    &- \frac{E^2}{2} \nabla \nabla \loss_{j} (\omega^{j}) \nabla \loss_{j} (\omega^{j}) + \frac{1}{2} \sum^{E-1}_{k=0} \nabla \nabla \loss_{jk} (\omega^{j}) \nabla \loss_{jk} (\omega^{j}) \\
    &= \sum^{E-1}_{k=0} \sum^{k-1}_{l<k} \nabla \nabla \loss_{jk} (\omega^{j}) \nabla (\loss_{jl} (\omega^{j}) - \loss_{j} (\omega^{j})) \nonumber \\
    &+ \frac{1}{2} \sum^{E-1}_{k=0} \nabla \nabla (\loss_{jk} (\omega^{j}) - \loss_{j} (\omega^{j})) \nabla (\loss_{jk} (\omega^{j}) - \loss_{j} (\omega^{j})) \\
    &= \sum^{E-1}_{k=0} \sum^{k-1}_{l<k} \nabla \nabla \loss_{jk} (\omega^{j}) \nabla (\loss_{jl} (\omega^{j}) - \loss_{j} (\omega^{j})) + \frac{1}{4} \sum^{E-1}_{k=0} \nabla \| \nabla (\loss_{jk} (\omega^{j}) - \loss_{j} (\omega^{j})) \|^2.
\end{align}
Also, the expectation over the discrepancy can be checked. Here we again use the fact that every sequence has its own reverse order and the average of $\xi^{j}$s in a forward order and a reverse order is the same for every sequence.
{\allowdisplaybreaks
\begin{align}
    \mathbb{E}(\xi^{j}(\omega^{j}) - \mathbb{E}(\xi^{j}(\omega^{j}))) &= \frac{1}{2} \sum^{E-1}_{k=0} \sum^{k-1}_{l \neq k} \nabla \nabla \loss_{jk} (\omega^{j}) \nabla (\loss_{jl} (\omega^{j}) - \loss_{j} (\omega^{j})) \nonumber \\
    &+ \frac{1}{2} \sum^{E-1}_{k=0} \nabla \nabla \loss_{jk} (\omega^{j}) \nabla \loss_{jk} (\omega^{j}) - \nabla \nabla \loss_{j} (\omega^{j}) \nabla \loss_{j} (\omega^{j}) \\
    &= \frac{1}{2} \sum^{E-1}_{k=0} \sum_{l \neq k} \nabla \nabla \loss_{jk} (\omega^{j}) \nabla (\loss_{jl} (\omega^{j}) - \loss_{j} (\omega^{j})) \nonumber \\
    &+ \frac{1}{2} \sum^{E-1}_{k=0} \nabla \nabla \loss_{jk} (\omega^{j}) (\nabla \loss_{jk} (\omega^{j}) - \nabla \loss_{j} (\omega^{j})) \\
    &= \frac{1}{2} \sum^{E-1}_{k=0} \sum^{E-1}_{l=0} \nabla \nabla \loss_{jk} (\omega^{j}) \nabla (\loss_{jl} (\omega^{j}) - \loss_{j} (\omega^{j})) = 0
\end{align}
}
and the expectation is zero, as it should be.

\paragraph{Local epochs and high-order terms.}
\label{proof:secondary}

Now we divide the number of local steps $E$ into two factors: the number of local epochs, $a$, and the number of local steps within a single local epoch, $K$. We now postulate that the number of updates within a local epoch, $K$, is small enough and the number of local epochs, $a$, is large enough to make third-order terms $O(a^3 K^3 \eta^3)$ considerable but $O(a K^3 \eta^3)$ negligible. We re-inspect the parameter updates within a local epoch to re-consider the dispersion term. Here, we assume that the local parameter updates are following the expectation of possible paths for ease of analysis.
\begin{align}
    \omega^{j}_K &= \omega^{j}_0 - K \eta \nabla \loss_{j} (\omega^{j}_0) + \frac{K^2 \eta^2}{4} \nabla \| \nabla \loss_{j} (\omega^{j}_0) \|^2 + O(K^3 \eta^3) \\
    \omega^{j}_{2K} &= \omega^{j}_K - K \eta \nabla \loss_{j} (\omega^{j}_K) + \frac{K^2 \eta^2}{4} \nabla \| \nabla \loss_{j} (\omega^{j}_K) \|^2 + O(K^3 \eta^3) \nonumber \\
    &= \omega^{j}_0 - 2K \eta \nabla \loss_{j} (\omega^{j}_0) + \frac{K^2 \eta^2}{2} \nabla \| \nabla \loss_{j} (\omega^{j}_0) \|^2 + \frac{K^2 \eta^2}{4} \nabla \| \nabla \loss_{j} (\omega^{j}_0) \|^2 + \frac{K^2 \eta^2}{4} \nabla \| \nabla \loss_{j} (\omega^{j}_K) \|^2 + O(K^3 \eta^3) \\
    \omega^{j}_{3K} &= \omega^{j}_{2K} - K \eta \nabla \loss_{j} (\omega^{j}_{2K}) + \frac{K^2 \eta^2}{4} \nabla \| \nabla \loss_{j} (\omega^{j}_{2K}) \|^2 + O(K^3 \eta^3) \nonumber \\
    &= \omega^{j}_K - 2K \eta \nabla \loss_{j} (\omega^{j}_K) + \frac{K^2 \eta^2}{2} \nabla \| \nabla \loss_{j} (\omega^{j}_K) \|^2 + \frac{K^2 \eta^2}{4} \nabla \| \nabla \loss_{j} (\omega^{j}_{K}) \|^2 + \frac{K^2 \eta^2}{4} \nabla \| \nabla \loss_{j} (\omega^{j}_{2K}) \|^2 + O(K^3 \eta^3) \nonumber \\
    &= \omega^{j}_0 - 3K \eta \nabla \loss_{j} (\omega^{j}_0) + 2 \cdot \frac{K^2 \eta^2}{2} \nabla \| \nabla \loss_{j} (\omega^{j}_0) \|^2 + \frac{K^2 \eta^2}{2} \nabla \| \nabla \loss_{j} (\omega^{j}_K) \|^2 \nonumber \\
    &+ \frac{K^2 \eta^2}{4} \nabla \| \nabla \loss_{j} (\omega^{j}_{0}) \|^2 + \frac{K^2 \eta^2}{4} \nabla \| \nabla \loss_{j} (\omega^{j}_{K}) \|^2 + \frac{K^2 \eta^2}{4} \nabla \| \nabla \loss_{j} (\omega^{j}_{2K}) \|^2 +O(K^3 \eta^3) \\
    \dots \nonumber \\
    \omega^{j}_{aK} &= \omega^{j}_0 - aK \eta \nabla \loss_{j} (\omega^{j}_0) + \sum^{a-1}_{l=0} (a-1-l) \cdot \frac{K^2 \eta^2}{2} \nabla \| \nabla \loss_{j} (\omega^{j}_{lK}) \|^2 + \frac{K^2 \eta^2}{4} \sum^{a-1}_{l=0} \nabla \| \nabla \loss_{j} (\omega^{j}_{lK}) \|^2 + O(a K^3 \eta^3)
    \label{eq:fromK}
\end{align}
The terms of $O(K^3 \eta^3)$ formed by $\dddot{\omega}^j (t)$ in the equations above were ignored. If we approximate $\omega^{j}_{lK} \approx \omega^{j}_0$ for $l \in \{0, 1, \dots, a-1\}$, the result above is the same as Equation \ref{eqa} we previously obtained.

After aggregation of local parameters, the dispersion term can also be obtained and rearranged following the order of their local epochs where they are introduced in. Here, we first define that the parameter is updated to $\tilde{\omega}_{lK}$ after the $l$-th local epoch if the parameter updates strictly follow the gradient flow of the original loss function.
\begin{equation}
    \tilde{\omega}_{lK} = \omega^{j}_0 - lK \eta \nabla \loss (\omega^{j}_0) + K^2 \eta^2 \sum^{l-1}_{i=0} (\frac{l - 1 - i}{2} + \frac{1}{4}) \nabla \| \nabla \loss (\tilde{\omega}_{iK}) \|^2 + O(a K^3 \eta^3)
\end{equation}
For ease of analysis and to induce a physical meaning from the second-order dispersion term, we use that $\frac{1}{m} \sum^{m-1}_{j=0} \omega^{j}_{lK} = \tilde{\omega}_{lK} + O(l^2 K^2 \eta^2)$ for $l \in \{0, 1, \dots, a-1\}$.
If we rearrange the second-order terms in the order of their local epochs, ignoring terms with an order higher than the third, we can get

\begin{eqnarray*}
    \omega^{j}_0\textrm{: } & & \frac{K^2 \eta^2}{m} (\frac{1}{4} + \frac{a-1}{2}) \sum^{m-1}_{j=0} \nabla \| \nabla \loss_{j} (\omega^{j}_0) - \nabla \loss (\tilde{\omega}_0) \|^2 \\
    \omega^{j}_K\textrm{: } & & \frac{K^2 \eta^2}{m} (\frac{1}{4} + \frac{a-2}{2}) \sum^{m-1}_{j=0} \nabla \| \nabla \loss_{j} (\omega^{j}_K) - \nabla \loss (\tilde{\omega}_K) \|^2 \\
    \dots \\
    \omega^{j}_{(a-1)K}\textrm{: } & & \frac{K^2 \eta^2}{m} (\frac{1}{4} + 0) \sum^{m-1}_{j=0} \nabla \| \nabla \loss_{j} (\omega^{j}_{(a-1)K}) - \nabla \loss (\tilde{\omega}_{(a-1)K}) \|^2.
    \label{eq:second-drift}
\end{eqnarray*}

About a dispersion term related to $\omega^{j}_{lK}$, we can again ignore subsidiary terms of an order higher than third to get the secondary dispersion term. To do so, we use the fact that $\omega^{j}_{lK} = \omega^{j} - lK \eta \nabla \loss_{j} (\omega^{j}_0) + O(l^2 K^2 \eta^2)$ and get
\begin{align}
    &\frac{K^2 \eta^2}{m} (\frac{1}{4} + \frac{a-1-l}{2}) \sum^{m-1}_{j=0} \nabla \| \nabla \loss_{j} (\omega^{j}_{lK}) - \nabla \loss (\tilde{\omega}_{lK}) \|^2 \nonumber \\
    &= \frac{K^2 \eta^2}{m} (\frac{1}{4} + \frac{a-1-l}{2}) \sum^{m-1}_{j=0} \nabla \| \nabla \loss_{j} (\omega^{j}_0) - \frac{lK\eta}{2} \nabla \| \nabla \loss_{j} (\omega_0) \|^2 - \nabla \loss (\omega_0) + \frac{lK\eta}{2} \nabla \| \nabla \loss (\omega_0) \|^2 + O(l^2 K^2 \eta^2) \|^2 \\
    &= \frac{K^2 \eta^2}{m} (\frac{1}{4} + \frac{a-1-l}{2}) \sum^{m-1}_{j=0} \nabla [\| \nabla \loss_{j} (\omega^{j}_0) - \nabla \loss (\omega_0) \|^2 \nonumber \\
    &- lK\eta \nabla \langle \nabla \loss_{j} (\omega^{j}_0) - \nabla \loss (\omega_0), \nabla (\| \nabla \loss_{j} (\omega_0) \|^2 - \| \nabla \loss (\omega_0) \|^2) \rangle + O(l^2 K^2 \eta^2)] \label{eqcoefficient}
\end{align}

We can reform the third-order term in a more comprehensive way. Beforehand, we define the difference of client gradients as the gradient $\nabla \zeta_j(\omega_0)$:
$$\nabla \zeta_j(\omega_0) := \nabla \loss_{j} (\omega_0) - \nabla \loss (\omega_0)$$ 
Then we get
\begin{align}
    &\frac{1}{m} \sum^{m-1}_{j=0} \langle \nabla \loss_{j} (\omega_0) - \nabla \loss (\omega_0), \nabla (\| \nabla \loss_{j} (\omega_0) \|^2 - \| \nabla \loss (\omega_0) \|^2) \rangle \nonumber \\
    &= \frac{1}{m} \sum^{m-1}_{j=0} \langle \nabla \loss_{j} (\omega_0) - \nabla \loss (\omega_0), \nabla (\| \nabla \loss_{j} (\omega_0) - \nabla \loss (\omega_0) + \nabla \loss (\omega_0) \|^2 - \| \nabla \loss (\omega_0) \|^2) \rangle \\
    &= \frac{1}{m} \sum^{m-1}_{j=0} [\langle \nabla \zeta_j(\omega_0), \nabla \| \nabla \zeta_j(\omega_0) \|^2 + 2 \nabla \langle \nabla \zeta_j(\omega_0), \nabla \loss (\omega_0) \rangle \rangle] \\
    &= \frac{1}{m} \sum^{m-1}_{j=0} \langle \nabla \zeta_j(\omega_0), \nabla \| \nabla \zeta_j(\omega_0) \|^2 \rangle + \frac{2}{m} \sum^{m-1}_{j=0} \langle \nabla \zeta_j(\omega_0), \nabla \nabla \zeta_j(\omega_0) \nabla \loss (\omega_0) \rangle \nonumber \\
    &+ \frac{2}{m} \sum^{m-1}_{j=0} \langle \nabla \zeta_j(\omega_0), \nabla \nabla \loss (\omega_0) \nabla \zeta_j(\omega_0) \rangle \label{eqhessvecuse} \\
    &= \frac{1}{m} \sum^{m-1}_{j=0} \langle \nabla \zeta_j(\omega_0), \nabla \| \nabla \zeta_j(\omega_0) \|^2 \rangle + \frac{2}{m} \sum^{m-1}_{j=0} \langle \nabla \nabla \zeta_j(\omega_0) \nabla \zeta_j(\omega_0), \nabla \loss (\omega_0) \rangle \nonumber \\
    &+ \frac{2}{m} \sum^{m-1}_{j=0} \langle \nabla \loss_{j} (\omega_0) - \nabla \loss (\omega_0), \nabla \nabla \loss (\omega_0) (\nabla \loss_{j} (\omega_0) - \nabla \loss (\omega_0)) \rangle\label{eqsymmhess} \\
    &= \frac{1}{m} \sum^{m-1}_{j=0} [\langle \nabla \loss_{j} (\omega_0) - \nabla \loss (\omega_0), \nabla \| \nabla \zeta_j(\omega_0) \|^2 \rangle + \langle \nabla \loss (\omega_0), \nabla \| \nabla \zeta_j(\omega_0) \|^2 \rangle] \nonumber \\
    &+ \frac{2}{m} \sum^{m-1}_{j=0} \langle \nabla \loss_{j} (\omega_0) - \nabla \loss (\omega_0), \nabla \nabla \loss (\omega_0) (\nabla \loss_{j} (\omega_0) - \nabla \loss (\omega_0)) \\
    &= \frac{1}{m} \sum^{m-1}_{j=0} \langle \nabla \loss_{j} (\omega_0), \nabla \| \nabla \loss_{j} (\omega_0) - \nabla \loss (\omega_0) \|^2 \rangle \nonumber \\
    &+ \frac{2}{m} \sum^{m-1}_{j=0} (\nabla \loss_{j} (\omega_0) - \nabla \loss (\omega_0))^\top \nabla \nabla \loss (\omega_0) (\nabla \loss_{j} (\omega_0) - \nabla \loss (\omega_0)) \label{eqthirdorder}
\end{align}

and Equation \ref{eqsymmhess} was deduced using a fact that the second-order partial derivative of the cost function is continuous to make our Hessian symmetric.

Using Equation \ref{eqthirdorder}, we can now get the secondary dispersion term. Summing the coefficient of Equation \ref{eqthirdorder} in Equation \ref{eqcoefficient} should be done first as
\begin{align}
\label{eqgetcoeff}
    -K^3 \eta^3 \sum^{a-1}_{l=0} l(\frac{1}{4} + \frac{a-1-l}{2}) = -\frac{a^3 K^3 \eta^3}{12} + O(a^2 K^3 \eta^3)
\end{align}
From Equation \ref{eqthirdorder} and Equation \ref{eqgetcoeff}, we can ignore subsidiary terms of a small magnitude and get an approximation of the secondary dispersion term to compose the modified loss $\tilde{\loss}_{FedAvg} (\omega)$, which is
\begin{align}
    \tilde{\loss}_{FedAvg}(\omega) &\approx \loss(\omega) - \frac{aK \eta}{4m} \sum_{j=0}^{m-1} \|\nabla \loss (\omega) - \nabla \loss_{j}(\omega)\|^2 \nonumber \\
    &+ \frac{a^2 K^2 \eta^2}{12m} \sum_{j=0}^{m-1} \langle \nabla \loss_{j} (\omega), \nabla \| \nabla \loss (\omega) - \nabla \loss_j (\omega) \|^2 \rangle \nonumber \\
    &+ \frac{a^2 K^2 \eta^2}{6m} \sum_{j=0}^{m-1} (\nabla \loss (\omega) - \nabla \loss_j (\omega))^\top \nabla \nabla \loss (\omega) (\nabla \loss (\omega) - \nabla \loss_j (\omega)).
    \label{beforearr}
\end{align}
The equation above holds when $1 / E^2 \ll \eta \ll 1 / E$, which makes $O(\eta)$ negligible but $O(E^2 \eta^2)$ considerable in the implicit regularizer of FedAvg. Also, on the number of local epochs, $a \gg 1$ should be met since terms of $O(a K^3 \eta^3)$ from $\dddot{\omega}^j$ were ignored in Equation \ref{eq:fromK} but terms of $O(a^3 K^3 \eta^3)$ were considered in Equation \ref{eqgetcoeff} to gain a rather `interpretable' equation for the modified loss. As a finite-difference approximation, since $\eta$ is assumed to be much smaller than $1 / E$, we can arrange and approximate Equation \ref{beforearr} as
% In experiments, the effect of the quadratic objective term dominated during convergence and the variance of client gradients decreased as the number of local steps $E$ increased.
\begin{equation}
    \tilde{\loss}_{FedAvg}(\omega) \approx \loss(\omega) - \frac{aK \eta}{4m} \sum_{j=0}^{m-1} \|\nabla \zeta_j (\omega - \frac{aK \eta}{3} \nabla \loss_j (\omega))\|^2 + \frac{a^2 K^2 \eta^2}{6m} \sum_{j=0}^{m-1} \nabla \zeta_j (\omega)^\top \nabla \nabla \loss (\omega) \nabla \zeta_j (\omega).
    \label{afterarr}
\end{equation}
Considering second-order terms, the implicit regularizer of FedAvg is comprised of two terms: $- \frac{aK \eta}{4m} \sum_{j=0}^{m-1} \|\nabla \zeta_j (\omega - \frac{aK \eta}{3} \nabla \loss_j (\omega))\|^2$ and $ \frac{a^2 K^2 \eta^2}{6m} \sum_{j=0}^{m-1} \nabla \zeta_j (\omega)^\top \nabla \nabla \loss (\omega) \nabla \zeta_j (\omega)$.
The former term follows the shape of the original dispersion term but it is about $\omega - \frac{a K \eta}{3} \nabla \loss_j (\omega)$ instead of $\omega$. 
Now the modified loss does not maximize the gradient variance at the exact current parameter $\omega$. Instead, it maximizes the gradient variance at $\omega - \frac{a K \eta}{3} \nabla \loss_j (\omega)$, which is a parameter already `drifted' apart from the direction of the global gradient. Such a change in the dispersion term reduces the effect of the dispersion term on increasing the gradient variance. Still, the change in the dispersion term does not mean that it removes the dispersion term. Rather, the transformation induces an additional bias to FedAvg, which can be detrimental to the convergence of FedAvg.
% When the former term is minimized, $\frac{a^2 K^2 \eta^2}{12m} \sum_{j=0}^{m-1} \langle \nabla \loss_{j} (\omega), \nabla \| \nabla \loss (\omega) - \nabla \loss_j (\omega) \|^2 \rangle$ is also minimized. The gradient of the client gradient variance becomes the opposite to $\nabla \loss_{j} (\omega)$, which is opposite to the direction of local parameter updates. It means that the former term aligns the gradient of the client gradient variance to the direction of local parameter updates and might induce FedAvg to accelerate the bias of the dispersion term.
% Such a transformation can induce an additional bias that can deflect the path of the parameter updates to a detrimental direction, which can be confirmed in empirical results.
On the other hand, the latter term is a quadratic objective function which minimizes the distance between the client gradient and the global gradient when the Hessian is positive-definite and all of its eigenvalues are positive values. When the loss Hessian has $k$-th eigenvalue and the corresponding eigenvector as $\lambda_k$ and $v_k$ respectively, then
\begin{equation}
\textstyle
    (\nabla \loss (\omega) - \nabla \loss_j (\omega))^\top \nabla \nabla \loss (\omega) (\nabla \loss (\omega) - \nabla \loss_j (\omega)) = \sum_k \lambda_k \langle v_k, \nabla \loss (\omega) - \nabla \loss_j (\omega) \rangle^2
\end{equation}
% On the other hand, when the former term is minimized, the gradient of the client gradient variance becomes counter to $\nabla \loss_{j} (\omega)$, which is opposite to the direction of local parameter updates. It means that the former term might induce FedAvg to accelerate the bias of the dispersion term.

The effects of the dispersion term and second-order terms of the implicit regularizer collide and which effect dominates during training depends on the number of local epochs. When $aK \eta$ is small, the effect of the dispersion term, which is in first-order, prevails during optimization and the parameter steers towards sharp minima. However, unlike the dispersion term of which the coefficient is linear in the number of local steps $E$, the order of $E$ in second-order terms is quadratic. When the number of local epochs increases, the magnitude of second-order terms grows faster than the linear rate of the dispersion term. It implies that second-order terms of the implicit regularizer amplify their influence when the number of local epochs increases and the parameter can steer towards relatively flatter minima.
However, though the second-order terms minimize the gradient variance and lead FedAvg to flat minima, the transformation of the dispersion term still adds a bias to FedAvg and leads FedAvg to a sub-optimal point. Despite the diminished effect of the dispersion term, the convergence of FedAvg can still be slow and the generalization of FedAvg can still be worsened.

% 왜 variance 줄이는지 설명

\section{A backward error analysis on SCAFFOLD}
\label{proof:scaffold}

For ease of analysis, we suppose that training is done under an ideal option noted in \citet{karimireddy2020scaffold}. During training of the $j$-th client at the $i$-th round, the parameter $\omega^{ij}$ is being updated based on the client control variate $c_{ij}$ and the server control variate $c_i$. First, we need to approximate $c_{ij}$ and $c_i$, which we will approximate as $\nabla \loss_j(\omega^{(i-1)j}_0)$ and $\nabla \loss(\omega^{(i-1)j}_0)$. Since the first round of SCAFFOLD starts with $c_{ij}$ and $c_i$ as zero, the first round is equivalent to FedAvg and we can say that $\omega^{ij}_0 = \omega^{(i-1)j}_0 - E \eta \nabla \loss(\omega^{(i-1)j}_0) + O(E^2 \eta^2)$ in the first round of SCAFFOLD. Also, it is possible to check that the assertion of $\omega^{ij}_0 = \omega^{(i-1)j}_0 - E \eta \nabla \loss(\omega^{(i-1)j}_0) + O(E^2 \eta^2)$ can be applied to all later rounds when it can be applied to the first round. Then we can obtain $\nabla \loss_j(\omega^{(i-1)j}_0)$
\begin{align}
    \nabla \loss_j(\omega^{(i-1)j}_0) &= \nabla \loss_j(\omega^{ij}_0 + E \eta \nabla \loss(\omega^{(i-1)j}_0) + O(E^2 \eta^2)) \nonumber \\
    &= \nabla \loss_j(\omega^{ij}_0 + E \eta \nabla \loss(\omega^{ij}_0 + E \eta \nabla \loss(\omega^{(i-1)j}_0)) + O(E^2 \eta^2)) \nonumber \\
    &= \nabla \loss_j(\omega^{ij}_0 + E\eta \nabla \loss(\omega^{ij}_0)) + O(E^2 \eta^2).
\end{align}
Since we will always multiply control variates with $\eta$ or $E \eta$, we ignore high-order terms and approximate $\loss_{j}$ as $\nabla \loss_j(\omega^{ij}_0 + E\eta \nabla \loss(\omega^{ij}_0))$. In the same way, we approximate $c_i$ as $\nabla \loss(\omega^{ij}_0 + E\eta \nabla \loss(\omega^{ij}_0))$. Now the discrete updates of the parameter $\omega^{ij}$ during $E$ steps can be expressed step-by-step.
\begin{align}
    \omega^{ij}_1 &= \omega^{ij}_0 - \eta (\nabla \loss_{j0}(\omega^{ij}_0) - c_{ij} + c_i) \\
    \omega^{ij}_2 &= \omega^{ij}_1 - \eta (\nabla \loss_{j1}(\omega^{ij}_0 - \eta(\nabla \loss_{j0}(\omega^{ij}_0) - c_{ij} + c_i)) - c_{ij} + c_i) \\
    \dots \nonumber \\
    \omega^{ij}_E &= \omega^{ij}_0 - E \eta (\nabla \loss_j(\omega^{ij}_0) - c_{ij} + c_i) + \frac{\eta^2}{2} \sum^{E-1}_{k=0} \sum^{k-1}_{l=0} \nabla \nabla \loss_{jk}(\omega^{ij}_0) (\nabla \loss_{jl}(\omega^{ij}_0) - c_{ij} + c_i) + O(E^3 \eta^3)
\end{align}
Neglecting $O(E^3 \eta^3)$ terms, the expectation of parameter $\omega^{ij}$ is expressed as
\begin{align}
    \mathbb{E}(\omega^{ij}_E) &= \omega^{ij}_0 - E \eta \nabla \loss(\omega^{ij}_0) - E^2 \eta^2 \nabla \nabla (\loss(\omega^{ij}_0) - \loss_j(\omega^{ij}_0)) \nabla \loss(\omega^{ij}_0) \nonumber \\
    &+ \frac{E^2 \eta^2}{4} \nabla (\| \nabla \loss_{j} (\omega^{ij}_0) \|^2 - \frac{1}{E^2} \sum^{E-1}_{k=0} \| \nabla \loss_{jk} (\omega^{ij}_0) \|^2) \nonumber \\
    & - \frac{E(E-1)}{4} \eta^2 \nabla \| \nabla \loss_{j} (\omega^{ij}_0) \|^2 + \frac{E(E-1)}{2} \eta^2 \nabla \nabla \loss_j(\omega^{ij}_0) \nabla \loss(\omega^{ij}_0) + O(E^3 \eta^3)
    \label{eqremoved}
\end{align}
After the $i$-th round, the client parameters are aggregated and form a parameter $\omega^i$.
\begin{align}
    \omega^i_E &= \omega^i_0 - E \eta \nabla \loss(\omega^i_0) \nonumber \\
    &+ \frac{E^2 \eta^2}{4} \nabla (\| \nabla \loss(\omega^i_0) \|^2 - \frac{1}{E} \| \nabla \loss(\omega^i_0) \|^2 \nonumber \\
    & + \frac{1}{mE} \sum^{m-1}_{j=0} \| \nabla \loss_j(\omega^i_0) \|^2 - \frac{1}{mE^2} \sum^{m-1}_{j=0} \sum^{E-1}_{k=0} \| \nabla \loss_{jk}(\omega^i_0) \|^2) + O(E^3 \eta^3)
\end{align}
Since all clients participate in every round, we can equate $\omega^i$ and $\omega$. Then the expectation value of global parameter $\omega$ is
\begin{align}
    \mathbb{E}(\omega_{E}) &= \omega_0 - E \eta \nabla \loss(\omega_0) \nonumber \\
    &+ \frac{E^2\eta^2}{4} \nabla (\| \nabla \loss(\omega_0) \|^2 - \frac{1}{E} \| \nabla \loss(\omega_0) \|^2 - \frac{1}{mE^2} \sum^{m-1}_{j=0} \sum^{E-1}_{k=0} \| \nabla \loss_j(\omega^i_0) - \nabla \loss_{jk}(\omega^i_0) \|^2) + O(E^3\eta^3)
\end{align}
Then the modified loss under SCAFFOLD is
\begin{align}
    \tilde{\loss}_{SCAFFOLD}(\omega) \approx \loss(\omega) + \frac{\eta}{4} \|\nabla \loss(\omega)\|^2 + \frac{\eta}{4mE} \sum^{m-1}_{j=0} \sum^{E-1}_{k=0} \| \nabla \loss_j(\omega) - \nabla \loss_{jk}(\omega) \|^2
\end{align}

Here the modified loss of SCAFFOLD does not contain the dispersion term. However, the control variates that remove the dispersion term are approximations from the global and client gradients of the previous round. While the gap between the control variates and real gradients was ignored altogether with high-order terms in Equation \ref{eqremoved}, if high-order terms are considered, the gap is not fully removed and the effect of such a gap can appear in practical situations. Therefore, the modified loss of SCAFFOLD is different from FedAvg without the dispersion term in the perspective of high-order terms.

\subsection{High-order terms of implicit regularizer of variance reduction methods}

We now focus on high-order terms of implicit regularizer of variance reduction methods that use stagnant gradients as control variates. For ease of analysis, we now assume that control variates are $c_{ij} = \nabla \loss_j (\omega^{ij}_0)$ and $c_i = \nabla \loss (\omega^{ij}_0)$, which is a setting similar to the one in MIME \cite{karimireddy2020mime}. To investigate the implicit regularization, we take a look at Equation \ref{eq:fromK} and fix it into an equation for a variance reduction method:
\begin{align}
    \omega^{ij}_K &= \omega^{ij}_0 - K \eta c_i + \frac{K^2 \eta^2}{4} \nabla \| \nabla \loss_{j} (\omega^{ij}_0) \|^2 - \frac{K^2 \eta^2}{2} \nabla \nabla \loss_j (\omega^{ij}_0) c_{ij} + \frac{K^2 \eta^2}{2} \nabla \nabla \loss_j (\omega^{ij}_0) c_i + O(K^3 \eta^3)
\end{align}
\begin{align}
    \omega^{ij}_{2K} &= \omega^{ij}_K - K \eta (\nabla \loss_j (\omega^{ij}_K) - c_{ij} + c_i) + \frac{K^2 \eta^2}{4} \nabla \| \nabla \loss_{j} (\omega^{ij}_K) \|^2 - \frac{K^2 \eta^2}{2} \nabla \nabla \loss_j (\omega^{ij}_K) c_{ij} + \frac{K^2 \eta^2}{2} \nabla \nabla \loss_j (\omega^{ij}_K) c_i + O(K^3 \eta^3) \nonumber \\
    &= \omega^{ij}_0 - 2K \eta c_i + K^2 \eta^2 \nabla \nabla \loss_{j} (\omega^{ij}_0) c_i + \frac{K^2 \eta^2}{4} \nabla \| \nabla \loss_{j} (\omega^{ij}_0) \|^2 + \frac{K^2 \eta^2}{4} \nabla \| \nabla \loss_{j} (\omega^{ij}_K) \|^2 \nonumber \\
    & - \frac{K^2 \eta^2}{2} \nabla \nabla \loss_j (\omega^{ij}_0) c_{ij} + \frac{K^2 \eta^2}{2} \nabla \nabla \loss_j (\omega^{ij}_0) c_i - \frac{K^2 \eta^2}{2} \nabla \nabla \loss_j (\omega^{ij}_K) c_{ij} + \frac{K^2 \eta^2}{2} \nabla \nabla \loss_j (\omega^{ij}_K) c_i + O(K^3 \eta^3)
\end{align}
\begin{align}
    \omega^{ij}_{3K} &= \omega^{ij}_{2K} - K \eta (\nabla \loss_{j} (\omega^{ij}_{2K}) - c_{ij} + c_i) + \frac{K^2 \eta^2}{4} \nabla \| \nabla \loss_{j} (\omega^{ij}_{2K}) \|^2 - \frac{K^2 \eta^2}{2} \nabla \nabla \loss_j (\omega^{ij}_{2K}) c_{ij} + \frac{K^2 \eta^2}{2} \nabla \nabla \loss_j (\omega^{ij}_{2K}) c_i + O(K^3 \eta^3) \nonumber \\
    &= \omega^{ij}_K - 2K \eta (\nabla \loss_j (\omega^{ij}_K) - c_{ij} + c_i) + K^2 \eta^2 \nabla \nabla \loss_{j} (\omega^{ij}_K) c_i + \frac{K^2 \eta^2}{4} \nabla \| \nabla \loss_{j} (\omega^{ij}_K) \|^2 + \frac{K^2 \eta^2}{4} \nabla \| \nabla \loss_{j} (\omega^{ij}_{2K}) \|^2 \nonumber \\
    & - \frac{K^2 \eta^2}{2} \nabla \nabla \loss_j (\omega^{ij}_K) c_{ij} + \frac{K^2 \eta^2}{2} \nabla \nabla \loss_j (\omega^{ij}_K) c_i - \frac{K^2 \eta^2}{2} \nabla \nabla \loss_j (\omega^{ij}_{2K}) c_{ij} + \frac{K^2 \eta^2}{2} \nabla \nabla \loss_j (\omega^{ij}_{2K}) c_i + O(K^3 \eta^3) \nonumber \\
    &= \omega^{ij}_0 - 3K \eta c_i + 2 K^2 \eta^2 \nabla \nabla \loss_j (\omega^{ij}_0) c_i + K^2 \eta^2 \nabla \nabla \loss_j (\omega^{ij}_K) c_i + \frac{K^2 \eta^2}{4} \nabla \| \nabla \loss_{j} (\omega^{ij}_K) \|^2 + \frac{K^2 \eta^2}{4} \nabla \| \nabla \loss_{j} (\omega^{ij}_{2K}) \|^2 \nonumber \\
    & - \frac{K^2 \eta^2}{2} \nabla \nabla \loss_j (\omega^{ij}_K) c_{ij} + \frac{K^2 \eta^2}{2} \nabla \nabla \loss_j (\omega^{ij}_K) c_i - \frac{K^2 \eta^2}{2} \nabla \nabla \loss_j (\omega^{ij}_{2K}) c_{ij} + \frac{K^2 \eta^2}{2} \nabla \nabla \loss_j (\omega^{ij}_{2K}) c_i \nonumber \\
    &+ \frac{K^2 \eta^2}{4} \nabla \| \nabla \loss_{j} (\omega^{ij}_0) \|^2 - \frac{K^2 \eta^2}{2} \nabla \nabla \loss_j (\omega^{ij}_0) c_{ij} + \frac{K^2 \eta^2}{2} \nabla \nabla \loss_j (\omega^{ij}_0) c_i + O(K^3 \eta^3)
\end{align}
\begin{align}
    \dots \nonumber \\
    \omega^{ij}_{aK} &= \omega^{ij}_0 - aK \eta \nabla \loss (\omega^{ij}_0) + \sum^{a-1}_{l=0} (a-1-l + \frac{1}{2}) \cdot K^2 \eta^2 \nabla \nabla \loss_j (\omega^{ij}_{lK}) \nabla \loss (\omega^{ij}_0) \nonumber \\
    &+ \frac{K^2 \eta^2}{2} \sum^{a-1}_{l=0} (\nabla \nabla \loss_{j} (\omega^{ij}_{lK}) \nabla \loss_{j} (\omega^{ij}_{lK}) - \nabla \nabla \loss_{j} (\omega^{ij}_{lK}) \nabla \loss_{j} (\omega^{ij}_0)) + O(a K^3 \eta^3)
\end{align}

After aggregation of local parameters, the parameter becomes
\begin{align}
    \omega^{i}_{aK} &= \omega^{i}_0 - aK \eta \nabla \loss (\omega^{ij}_0) + \sum^{a-1}_{l=0} (a-1-l + \frac{1}{2}) \cdot K^2 \eta^2 \nabla \nabla \loss (\omega^{ij}_{lK}) \nabla \loss (\omega^{ij}_0) \nonumber \\
    &+ \frac{K^2 \eta^2}{2m} \sum^{m-1}_{j=0} \sum^{a-1}_{l=0} (\nabla \nabla \loss_{j} (\omega^{ij}_{lK}) \nabla \loss_{j} (\omega^{ij}_{lK}) - \nabla \nabla \loss_{j} (\omega^{ij}_{lK}) \nabla \loss_{j} (\omega^{ij}_0)) + O(a K^3 \eta^3)
\end{align}

If the parameter above is compared to the parameter updated from the gradient flow of the original loss function, which is 
\begin{align}
\tilde{\omega}^{i}_{aK} &= \omega^{i}_0 - aK \eta \nabla \loss_{i} (\omega^{j}_0) + \sum^{a-1}_{l=0} (a-1-l + \frac{1}{2}) \cdot K^2 \eta^2 \nabla \nabla \loss (\tilde{\omega}^{i}_{lK}) \nabla \loss (\tilde{\omega}^{i}_{lK}) + O(a K^3 \eta^3)
\end{align}
and the gap between those parameters is
\begin{align}
    \omega^{i}_{aK} - \tilde{\omega}^{i}_{aK} &\approx \sum^{a-1}_{l=0} (a-1-l + \frac{1}{2}) \cdot K^2 \eta^2 \nabla \nabla \loss (\tilde{\omega}^{i}_{lK}) (\nabla \loss (\omega^{i}_0) - \nabla \loss (\tilde{\omega}^{i}_{lK})) \nonumber \\
    &+ \frac{K^2 \eta^2}{2m} \sum^{m-1}_{j=0} \sum^{a-1}_{l=0} (\nabla \nabla \loss_{j} (\tilde{\omega}^{i}_{lK}) \nabla \loss_{j} (\omega^{i}_{lK}) - \nabla \nabla \loss_{j} (\tilde{\omega}^{i}_{lK}) \nabla \loss_{j} (\omega^{i}_0)) \\
    &\approx \sum^{a-1}_{l=0} (a-1-l + \frac{1}{2}) \cdot l K^3 \eta^3 \nabla \nabla \loss (\tilde{\omega}^{i}_{lK}) \nabla \nabla \loss (\tilde{\omega}^{i}_{lK}) \nabla \loss (\tilde{\omega}^{i}_{lK}) \nonumber \\
    &- \sum^{m-1}_{j=0} \sum^{a-1}_{l=0} \frac{l K^3 \eta^3}{2m} \nabla \nabla \loss_{j} (\tilde{\omega}^{i}_{lK}) \nabla \nabla \loss_{j} (\tilde{\omega}^{i}_{lK}) \nabla \loss_{j} (\tilde{\omega}^{i}_{lK})
\end{align}
The equation above is valid when $1 / E^2 \ll \eta \ll 1 / E$, which makes $O(\eta)$ negligible but $O(E^2 \eta^2)$ considerable in the implicit regularizer of FedAvg. Also, on the number of local epochs, $a \gg 1$ should be met since terms of $O(a K^3 \eta^3)$ from $\dddot{\omega}^j$ were ignored but terms of $O(a^3 K^3 \eta^3)$ were considered.

Although the equation above does not have a clear form that tells something about the modified loss, it tells that SCAFFOLD deviates from the path of the original gradient flow in high-order terms of the implicit regularizer. Such a deviation will become severe if the number of local steps and the size of a learning rate becomes larger, or the eigenvalues of Hessian becomes larger.

\section{Fisher information and Hessian}
\label{connection}

% FedAvg is almost identical to local SGD in a sense that both algorithms distribute samples to workers to train them in parallel and aggregate the trained parameters from workers by averaging them \cite{stich2018local}. Meanwhile, there was another attempt to analyze the dynamics of the gradient flow in the case of local SGD \cite{gu2023why}. Under the assumption on a small magnitude of the learning rate, \citeauthor{gu2023why} captures the long-term behaviour of local SGD and assert that local SGD has a stronger implicit bias towards flat minima than SGD. Thanks to the similarity of local SGD and FedAvg, it is also possible to analyze the gradient flow of local SGD based on our analysis of FedAvg. Instead, rather than capturing the long-term behaviour of local SGD, we obtained the behaviour in a relatively short-term where the product of the learning rate and the number of local steps is not large. One resemblance of the analysis is that we also assume that the optimized problem is a maximum likelihood estimation problem.

Now we assume that the current parameter $\omega$ is close to the optima and the outputs of a model in the current parameter are almost identical to the ground-truth. Also, as mentioned before, we assume that the current problem is a maximum likelihood estimation problem, for example, where cross-entropy is a minimized loss function. Under such assumptions, we can approximate the Hessian of a loss as Fisher information matrix \cite{kirkpatrick2017overcoming}. We can take the expectation of Hessian as below to obtain the approximation as below,

\begin{align}
    \mathcal{H}(\omega) &= \nabla \nabla \loss(\omega) = \mathbb{E}_{x,y} [\nabla \nabla_{\omega} (-\log p(y|x, \omega))] \approx \mathbb{E}_{x,y|\omega} [\nabla \nabla_{\omega} (-\log p(y|x, \omega))] \\ 
    &= \mathbb{E}_{x,y|\omega} [-\frac{\nabla \nabla_{\omega} p(y|x, \omega)}{p(y|x, \omega)} + \frac{\nabla_\omega p(y|x, \omega) \nabla_\omega p(y|x, \omega)^\top}{p(y|x, \omega)^2}] \\
    &= \mathbb{E}_{x,y|\omega} [-\frac{\nabla \nabla_{\omega} p(y|x, \omega)}{p(y|x, \omega)} + \nabla_\omega \log p(y|x, \omega) \nabla_\omega \log p(y|x, \omega)^\top] \\
    &= \mathbb{E}_{x,y|\omega} [\nabla_\omega \log p(y|x, \omega) \nabla_\omega \log p(y|x, \omega)^\top] \\
    &\approx \frac{1}{N} \sum^N_{i=1} \nabla_\omega \log p(y_i|x_i, \omega) \nabla_\omega \log p(y_i|x_i, \omega)^\top = \frac{1}{N} \sum^N_{i=1} \nabla \tilde{\loss}_i(\omega) \nabla \tilde{\loss}_i(\omega)^\top
\end{align}

when the number of samples in the dataset is $N$ and the gradient of the $i$-th sample is $\nabla \tilde{\loss}_i(\omega)$. The approximation in Equation 70 is based on the assumption that the current parameter $\omega$ is close to the optima, and Equation 73 is because $\mathbb{E}_{x, y|\omega}[\frac{\nabla \nabla_{\omega} p(y|x, \omega)}{p(y|x, \omega)}] = \nabla \nabla_{\omega} \int_{x, y|\omega} p(y|x, \omega) = 0$. Hessian, $\mathcal{H}(\omega) = \nabla \nabla \loss(\omega)$, can be approximated as the average of matrix products between the gradients of the sample and their transposed ones, which is called \textit{Fisher information matrix}. While Hessian can be approximated as such, considering that $\nabla \loss(\omega)$ is nearly zero when $\omega$ is close to the optima, it is possible to notice that the trace of Fisher information can be approximated as the variance of gradients from samples. 

\begin{equation}
    \mathrm{tr}(\nabla \nabla \loss(\omega)) \approx \frac{1}{N} \sum^N_{i=1} \| \nabla \tilde{\loss}_i(\omega) - \nabla \loss(\omega) \|^2
    \label{eqtrace}
\end{equation}

Based on this approximation, we obtained the results on the dispersion term and the secondary dispersion term.

\section{FedAvg and sharp minima}

\begin{figure*}[hbtp]
    \centering
    \subfigure[]{
        \includegraphics[width=.25\textwidth]{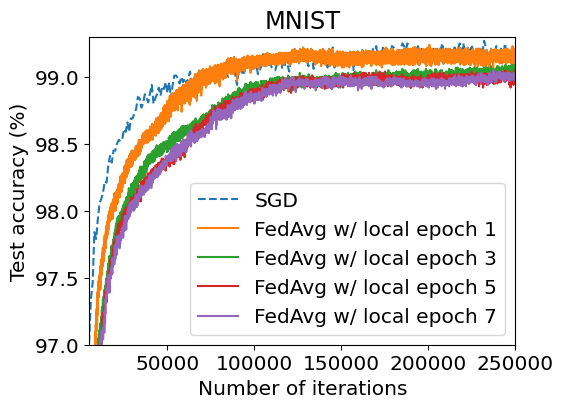}
    }
    \hfill
    \subfigure[]{
        \includegraphics[width=.23\textwidth]{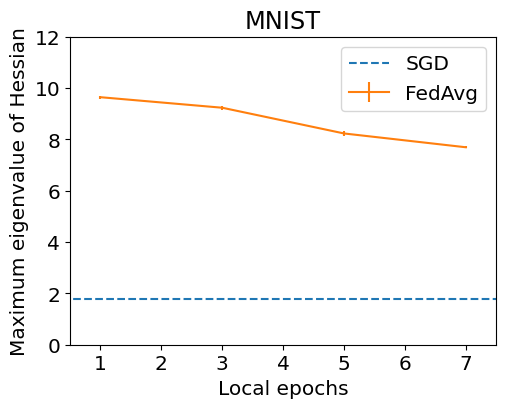}
    }
    \hfill
    \subfigure[]{
        \includegraphics[width=.25\textwidth]{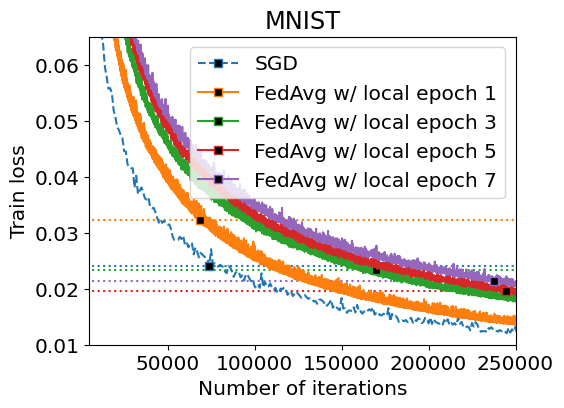}
    }
    \subfigure[]{
        \includegraphics[width=.25\textwidth]{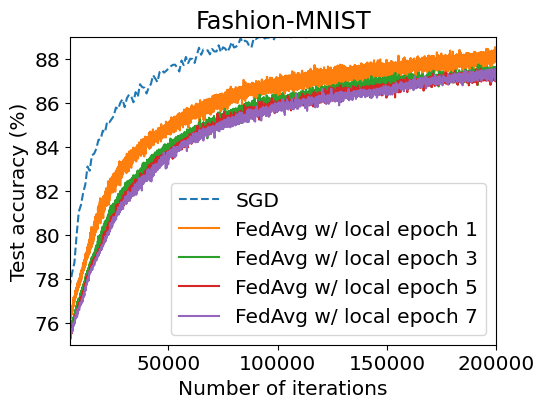}
    }
    \hfill
    \subfigure[]{
        \includegraphics[width=.23\textwidth]{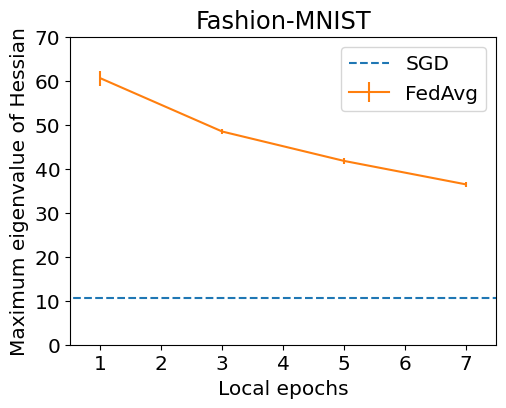}
    }
    \hfill
    \subfigure[]{
        \includegraphics[width=.25\textwidth]{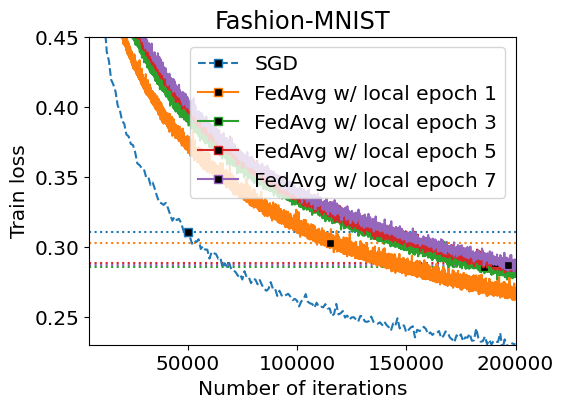}
    }
    \subfigure[]{
        \includegraphics[width=.25\textwidth]{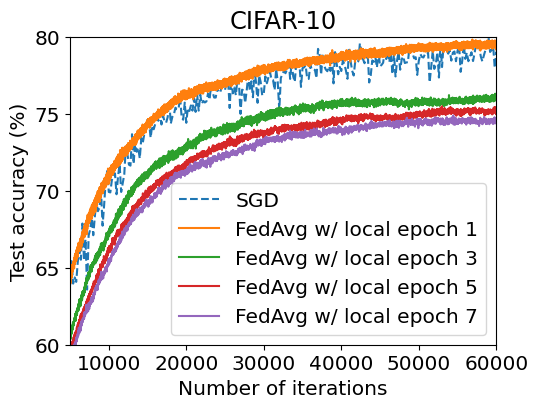}
    }
    \hfill
    \subfigure[]{
        \includegraphics[width=.23\textwidth]{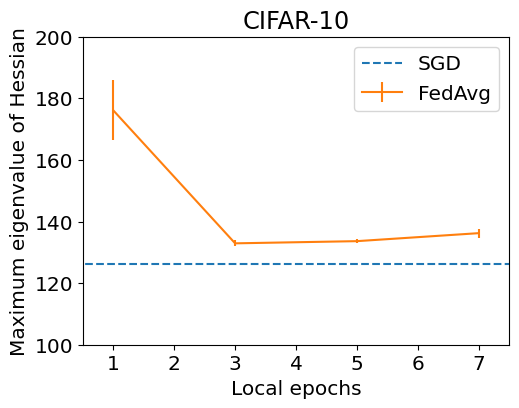}
    }
    \hfill
    \subfigure[]{
        \includegraphics[width=.25\textwidth]{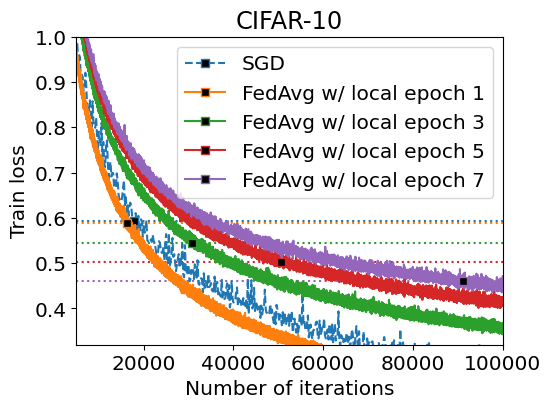}
    }
    \caption{Test accuracy, maximum eigenvalue, and train loss at the end of training of FedAvg and SGD on MNIST, Fashion-MNIST, and CIFAR-10. As the number of local epochs increases, maximum eigenvalue of FedAvg decrease but it did not lead to generalization of FedAvg.}
    \label{FigEigen}
\end{figure*}

\paragraph{Experimental settings.}

We ran experiments on MNIST, Fashion-MNIST, and CIFAR-10 for 30,000 epochs, while the step size was 0.01 for all datasets, and the batch size was set 60 on MNIST, Fashion-MNIST, 300 on CIFAR-10 both for FedAvg and SGD. The step size was set larger than before to enlarge the magnitude of $E \eta$, but the batch size was set large on CIFAR-10 because SGD was unstable at larger learning rates. All experiments were conducted on non-IID data of a Dirichlet distribution with parameter 0.2.

\paragraph{Dispersion term leads to sharp minima but second-order terms do not.}
In order to inspect the influence of the number of local steps, we ran experiments with different values of local epochs, which are 1, 3, 5, and 7. As shown in Figure \ref{FigEigen}, in the case of FedAvg with the local epoch of 1, the maximum eigenvalue of the loss Hessian at the converged point was larger than the one of SGD. This shows that the dispersion term increases the sharpness of the loss surface. On the other hand, when the number of local epochs increased to a value larger than 1, the maximum eigenvalue of the loss Hessian decreased and FedAvg converged to relatively flatter minima. 
% As the number of local steps increased, FedAvg overall converged to a relatively flatter minimum.
% It is due to the increased secondary dispersion terms, which grow faster than a linear rate when the number of local steps grows. 
This is because the high-order terms of the implicit regularizer increased its influence and reduced the sharpness of Hessian as the number of local epochs increased. 
% One thing to note is that such a phenomenon occurs dramatically on CIFAR-10 and the maximum eigenvalue of FedAvg becomes smaller than the one of SGD when the number of local epoch grows. 
% Such a result suggests that FedAvg with sufficient local steps might aid generalization of the model. 
While this also partially accords with \citet{gu2023why} and \citet{lin2019don}, it differs in that the sharpness was still consistently larger in FedAvg than SGD.
% However, one thing to note in the case of CIFAR-10 is that the maximum eigenvalue slightly increased when the number of local epochs became larger than 3, which implies that terms of higher orders in implicit regularizer also affect the convergence behaviour. 

\paragraph{Flat minima do not necessarily lead to generalization.}
The results in Figure \ref{FigEigen} show that the train loss around convergence becomes lower when local steps grow less. This result aligns with \citet{li2020convergence} in that FedAvg converges to a sub-optimal point when the number of local steps increases. An important question would be whether the test accuracies are also subpar in those converged `sub-optimal' points, though the parameter converges to flatter minima when the number of local steps is larger.
Firstly, the test accuracy at the converged point was larger when the number of local epochs was smaller, meaning that the model did not generalize when there were more local epochs.
As an additional metric, we measured the train losses of FedAvg and SGD when they exceeded the same test accuracies of 99.05 on MNIST, 87.5 on Fashion-MNIST, 75.2 on CIFAR-10. The marks and dotted lines in Figure \ref{FigEigen}(c), (f), (i) denote those train losses. When the number of local epochs became larger than 1, those measured train losses did not increase but instead decreased, which means that 
% the model needed more efforts to fit to trained data when the number of local steps increased. T
the model did not generalize even though the parameter inclined towards flat minima. This result suggests that flat minima do not always lead to generalization. Although high-order terms of the implicit regularizer lead FedAvg to flat minima, the entire implicit regularizer of FedAvg as a whole deflects the path of the parameter updates into a detrimental direction and can lead to worse generalization.
% It is the bias that matters for generalization, and bias of FedAvg on non-IID data can lead to worse generalization unlike FedAvg on IID data \cite{gu2023why, lin2019don}.

% However on CIFAR-10, when local epoch was larger than 5, the maximum eigenvalue at the converged parameter slightly increased. This implies that terms of higher orders in the implicit regularization can increase the trace of the loss Hessian. But judging by the result on maximum eigenvalues,  their effect was much smaller compared to the effect of the secondary dispersion terms. On MNIST and Fashion-MNIST, when the number of local steps increased more, the gradient variance and maximum eigenvalue decreased even more, which implies that the effect of high-order dispersion terms increased further. However on CIFAR-10, the gradient variance and maximum eigenvalue did not decrease and remained similar when the number of local steps increased. This shows that the effect of high-order dispersion term does not always increase with the number of local steps and FedAvg might still converge to sharper minima.
% CIFAR-10

\section{Algorithms for experiments}
\label{algorithm:experiment}

\paragraph{FedAvg}

This is an algorithm of FedAvg with local epochs of $a$ and the total communication rounds of $R$. The number of local steps within a single local epoch of $j$-th client is $K_j$. $K_j$ was asssumed to be the same as $K$ for all clients in our previous analyses. The aggregated parameter in the $i$-th round is $\omega^i$, while the parameter of $j$-th client in the $i$-th round is $\omega^{ij}$.

\begin{algorithm}[h]
    \caption{FedAvg}
    \label{algorithm:FedAvg}
    \begin{algorithmic}
        \STATE {\bfseries Input:} initial parameter $\omega^0$, learning rate $\eta$, numeber of local epochs $a$
        \FOR{communication round $i = 1$ \textbf{to} $R$}
            \STATE Sample clients $\mathcal{S}$, distribute $\omega^{ij} \leftarrow \omega^{i-1}$ to clients
            \FOR{client $j \in \mathcal{S}$ \textbf{in parallel}}
                \FOR{local epoch $l = 1$ \textbf{to} $a$}
                    \FOR{local step $k = 1$ \textbf{to} $K_j$}
                        \STATE $\omega^{ij} = \omega^{ij} - \eta \nabla \loss_{jk} (\omega^{ij})$
                    \ENDFOR
                \ENDFOR
            \ENDFOR
            \STATE $\omega^i = \frac{1}{|\mathcal{S}|} \sum^{\mathcal{S}}_{j=1} \omega^{ij}$
        \ENDFOR
    \end{algorithmic}
\end{algorithm}

\paragraph{FedAvg without the dispersion term}

We first calculated the average of client gradients $\nabla \loss(\omega^{i-1})$ and $\nabla \|\nabla \loss_i(\omega^{i-1})\|^2$ in the server and sent them to the clients. In the clients, we calculated the average of its mini-batch gradients $\nabla \loss_{j}(\omega^{i-1})$. We then obtained $\frac{E^2 \eta^2}{4} (\nabla \| \nabla \loss_{j}(\omega^{i-1})\|^2 - \nabla \|\nabla \loss(\omega^{i-1})\|^2)$ and subtracted it from the parameter of the client. The modified loss of the aggregated parameter now is $\tilde{\loss}_{modified}(\omega) = \loss(\omega) + \frac{\eta}{4mE} \sum_{j=0}^{m-1} \sum_{k=0}^{E-1} \|\nabla \loss_{jk}(\omega)\|^2$.

\begin{algorithm}[h]
    \caption{FedAvg without the dispersion term}
    \label{algorithm:FedAvgwo}
    \begin{algorithmic}
        \STATE {\bfseries Input:} initial parameter $\omega^0$, learning rate $\eta$, number of local epochs $a$
        \FOR{communication round $i = 1$ \textbf{to} $R$}
            \STATE Sample clients $\mathcal{S}$, distribute $\omega^{ij} \leftarrow \omega^{i-1}$ to clients
            \STATE Obtain $\nabla \loss(\omega^{i-1}) = \frac{1}{|\mathcal{S}|} \sum^{\mathcal{S}}_{j=1} \nabla \loss_j (\omega^{i-1})$, $\nabla \| \nabla \loss(\omega^{i-1}) \|^2$ and send them to clients
            \FOR{client $j \in \mathcal{S}$ \textbf{in parallel}}
                \STATE Obtain $\nabla \loss_j (\omega^{i-1}) = \frac{1}{K_j} \sum^{K_j}_{k=1} \nabla \loss_{jk} (\omega^{i-1})$ and $\nabla \| \nabla \loss_j (\omega^{i-1}) \|^2$
                \FOR{local epoch $l = 1$ \textbf{to} $a$}
                    \FOR{local step $k = 1$ \textbf{to} $K_j$}
                        \STATE $\omega^{ij} = \omega^{ij} - \eta \nabla \loss_{jk} (\omega^{ij})$
                    \ENDFOR
                \ENDFOR
                \STATE $\omega^{ij} = \omega^{ij} - \frac{a^2 K^2_j \eta^2}{4} \nabla \| \nabla \loss_j (\omega^{i-1}) \|^2 + \frac{a^2 K^2_j \eta^2}{4} \nabla \| \nabla \loss (\omega^{i-1}) \|^2$
            \ENDFOR
            \STATE $\omega^i = \frac{1}{|\mathcal{S}|} \sum^{\mathcal{S}}_{j=1} \omega^{ij}$
        \ENDFOR
    \end{algorithmic}
\end{algorithm}

\paragraph{$\varepsilon$ of FedSAM}

About FedSAM, we modified the value of $\varepsilon$ in Equation \ref{eqsam} for stability and performance. Unlike in original works where $\varepsilon$ was inversely proportional to the gradient norm \cite{foret2021sharpnessaware, qu2022generalized, caldarola2022fedasam}, $\varepsilon$ in our algorithm is inversely proportional to the square-root of the pseudo-gradient-norm. Our modification stabilized the variance of batch gradients, the value of $\varepsilon$, and performance.

\begin{algorithm}[h]
    \caption{Obtaining $\varepsilon$ for modified FedSAM}
    \begin{algorithmic}
        \STATE $norm = 0$
        \FOR{$grad$ in $\nabla \loss_{jk}(\omega)$}
            \STATE $norm = norm + \|grad\|_2$
        \ENDFOR
        
        \RETURN return $\varepsilon = 0.01 / \sqrt{norm}$
    \end{algorithmic}
\end{algorithm}

\section{Details on the experimental settings}

\subsection{Details on the model architecture}

A simple CNN model used in empirical analysis on the dispersion term for MNIST consists of 2 convolutional layers with 10 and 20 5 $\times$ 5 filters and ReLU activation function, fully-connected layers of 50 neurons, and a softmax layer.

A more complex CNN model used in experiments for FEMNIST and empirical analysis on FedSAM consists of 2 convolutional layers with 32 and 64 7 $\times$ 7 filters and ReLU activation function, and a softmax layer. 

A model used in experiments on high-order terms of implicit regularization consists of a single convolutional layer of 64 5 $\times$ 5 kernels, two 64-channel-2-stride basic convolutional blocks with a skip connection and instance normalization, two fully-connected layers with 384, 192 neurons and a softmax layer. A basic convolutional block consists of a convolutional layer of 64 5 $\times$ 5 filters with an instance normalization, a convolutional layer of 64 3 $\times$ 3 filters with an instance normalization, and a convolutional layer of 64 1 $\times$ 1 filters with an instance normalization. A skip connection is applied to the last convolutional layer of a basic block.

\subsection{Empirical analysis on the dispersion term}

We ran experiments with a simple CNN model described above on MNIST and FEMNIST. Experiments were done on a non-IID environment of Dirichlet distribution with parameter 0.2, except for FEMNIST, which is naturally non-IID. The batch size was 30 for MNIST and 100 for FEMNIST. The batch size was set the same both for FedAvg and SGD. We used a normal SGD optimizer with a small learning rate of 0.001 with no momentum and learning rate decay. The models were trained with 5 local epochs for 1000 rounds for MNIST, Fashion-MNIST, and for 2000 rounds for FEMNIST.

\subsection{Empirical analysis on FedSAM}

We have done experiments on MNIST and Fashion-MNIST with full client participation. Experiments were done on a non-IID environment of Dirichlet distribution with parameter 0.2. For stability of experiments, we switched the model to a more complex CNN model described above. The learning rate was 0.001 and the batch size was 60. The model was trained with 5 local epochs for 300 rounds on MNIST, with 5 local epochs for 500 rounds on Fashion-MNIST.

\subsection{Empirical analysis on the secondary dispersion term}

We ran experiments on CIFAR-10. Experiments were done on a non-IID environment and we use a Dirichlet distribution with parameter 0.05, which is more extreme than the previous experiment. 100 clients have been trained with a learning rate of 0.001 with no momentum and learning rate decay for 5000 rounds, consisting of 3 local epochs with the batch size of 300.

% \bibliography{aaai25}

\end{document}